\documentclass[11pt]{article}

\usepackage{acl}

\usepackage{times}
\usepackage{latexsym}
\usepackage{amsmath}
\usepackage[T1]{fontenc}
\usepackage[utf8]{inputenc}
\usepackage{microtype}
\usepackage{multirow}
\usepackage{caption}
\usepackage{enumitem}
\usepackage{booktabs}   
\usepackage{graphicx}   
\usepackage[table]{xcolor}
\usepackage{colortbl}
\usepackage{amssymb}
\usepackage{subfigure}
\usepackage{todonotes}

\colorlet{VanillaColor}{yellow!5!white}
\colorlet{RAGColor}{green!5!white}
\colorlet{MemoryColor}{cyan!8!white}
\colorlet{StructuralMemoryColor}{purple!8!white}

\title{StructMem: Structured Memory for Long-Horizon Behavior in LLMs}

\author{
  Buqiang Xu\textsuperscript{1}\thanks{ \ \ Equal contribution.},
  Yijun Chen\textsuperscript{1}\footnotemark[1],
  Jizhan Fang\textsuperscript{1},
  Ruobin Zhong\textsuperscript{1}, \\
  \textbf{Yunzhi Yao\textsuperscript{1}},
  \textbf{Yuqi Zhu\textsuperscript{1,3}},
  \textbf{Lun Du\textsuperscript{2,3}}, 
  \textbf{Shumin Deng\textsuperscript{1}}\thanks{ \ \ Corresponding author.}
  \\
  \\
  \textsuperscript{1}Zhejiang University
  \quad
  \textsuperscript{2}Ant Group
  \\
  \textsuperscript{3}Zhejiang University - Ant Group Joint Laboratory of Knowledge Graph \\
}

\begin{document}

\maketitle

\begin{abstract}
Long-term conversational agents need memory systems that capture relationships between events, not merely isolated facts, to support temporal reasoning and multi-hop question answering. Current approaches face a fundamental trade-off: flat memory is efficient but fails to model relational structure, while graph-based memory enables structured reasoning at the cost of expensive and fragile construction. To address these issues, we propose \textbf{StructMem}, a structure-enriched hierarchical memory framework that preserves event-level bindings and induces cross-event connections. By temporally anchoring dual perspectives and performing periodic semantic consolidation, StructMem improves temporal reasoning and multi-hop performance on \texttt{LoCoMo}, while substantially reducing token usage, API calls, and runtime compared to prior memory systems\footnote{\url{https://github.com/zjunlp/LightMem}}.

\end{abstract}

\section{Introduction}








Persistent memory systems are essential for language model agents to maintain coherence in long-term interactions~\cite{park2023generative}. 
Beyond factual recall, long-horizon dialogue requires reasoning over temporal dependencies, causal chains, and multi-hop relationships across turns~\cite{weller2025theoreticallimitationsembeddingbasedretrieval,huang2025licomemory,maharana2024evaluating,wu2024longmemeval,yang2018hotpotqa}. This necessitates memory representations that organize events into temporally grounded and relational structures~\cite{kwiatkowski2019natural}. 

\begin{figure}[!t]
    \centering
    \includegraphics[width=0.99\linewidth]{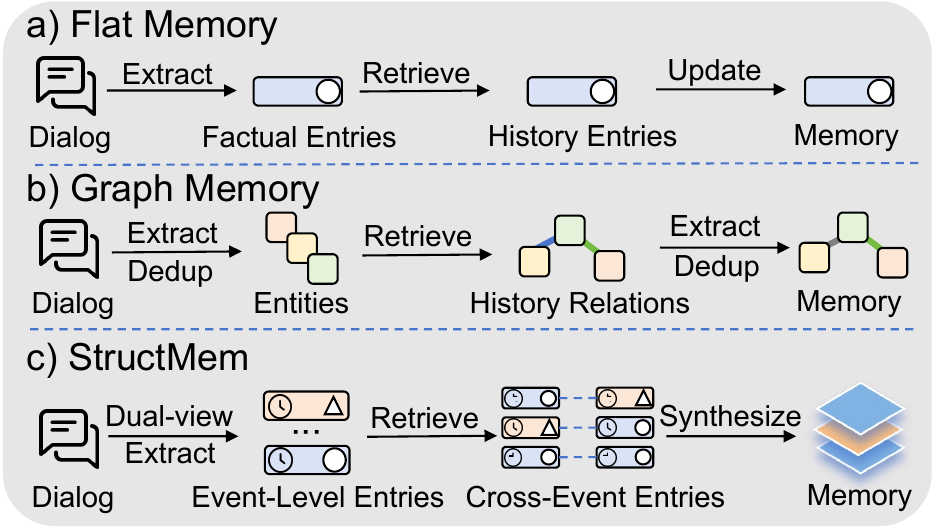}
    \caption{Three paradigms of Memory systems.}
    \vspace{-5mm}
    \label{fig:intro}
\end{figure}

Existing memory systems largely fall into two paradigms, flat memory and graph memory, exhibit a trade-off between efficiency and structured reasoning, as illustrated in Figure~\ref{fig:intro}. 
Specifically, flat memory systems~\cite{fang2025lightmem,zhong2024memorybank,packer2023memgpt} store facts or summaries as independent units, but fail to preserve cross-event relations, causing retrieval over long histories to degrade into shallow similarity matching~\cite{Liu2023LostIT,zhuang2025linearrag}. 
Graph-based systems~\cite{chhikara2025mem0,rasmussen2025zep} recover relational structure via entity–relation extraction, but incur high construction cost, require cascaded inference~\cite{edge2024local}, and are vulnerable to error accumulation from noisy extractions~\cite{zhuang2025linearrag}.
We argue that these limitations arise from an inappropriate memory unit. Rather than isolated facts or triplets, the fundamental unit of conversational memory should be a \emph{temporally grounded relational event}, which preserves causal and interpersonal context without imposing rigid schemas.

Based on this insight, we propose \textbf{StructMem}, a hierarchical memory framework built around event-centric representations. This abstraction preserves both what happened and how events relate across agents and time, while avoiding explicit schema design, entity resolution, and symbolic graph traversal. 
Specifically, at the event level, StructMem constructs structured episodes through dual-perspective extraction, capturing both event content and interactional relations within temporal context. At the cross-event level, it performs periodic consolidation over semantically related events, exploiting temporal locality to efficiently induce higher-level relational structure. Experiments on \texttt{LoCoMo} show that StructMem improves long-horizon reasoning while significantly reducing computational overhead.

\section{Related Work} \label{Related Work}
Long-term memory serves as the cognitive foundation for agents to maintain persona consistency and perform reasoning across extended horizons~\cite{maharana2024evaluating,wu2024longmemeval,dong2025towards,huang2025licomemory}.
 
Early approaches addressed the context window limitation by externalizing history into flat vector databases~\cite{park2023generative,packer2023memgpt,zhong2024memorybank}. While efficient for semantic matching, this paradigm fundamentally treats interaction history as an unordered bag of propositions, severing the temporal progression, causal dependencies, and relational substrate that bind events into coherent narratives~\cite{gao2023retrieval, Liu2023LostIT}. This flat representation leads to fragmented retrieval where isolated facts are returned without the contextual scaffolding necessary for complex reasoning~\cite{weller2025theoreticallimitationsembeddingbasedretrieval,li2025sculptorempoweringllmscognitive}. Recent work has explored enhanced retrieval strategies through reflective reasoning and closed-loop control mechanisms~\cite{du2025memr3memoryretrievalreflective}, yet these improvements still operate within the fundamental constraints of flat representations. Even with extended context windows, flat memory systems suffer from the Lost-in-the-Middle phenomenon~\cite{Liu2023LostIT}, where attention mechanisms degrade in ultra-long sequences, ultimately reducing multi-hop reasoning to superficial similarity search over disconnected facts~\cite{zhuang2025linearrag}.

To bridge this reasoning gap, the field has increasingly pivoted towards structure-enriched architectures, particularly those leveraging Knowledge Graphs. Static graph approaches, such as Microsoft GraphRAG~\cite{edge2024local} and HippoRAG~\cite{gutierrez2025rag}, employ hierarchical community detection and Personalized PageRank to facilitate global sense-making and multi-hop traversal. Concurrently, dynamic memory systems tailored for agents, such as $\text{Mem0}^\text{g}$~\cite{chhikara2025mem0} and Zep~\cite{rasmussen2025zep}, have introduced evolving schemas to capture the fluidity of user interactions. Recent advances further explore trainable graph representations~\cite{xia2025experience} and lightweight hierarchical graphs with entity-relation indexing~\cite{huang2025licomemory}, demonstrating substantial improvements in multi-agent collaboration~\cite{zhang2025g} and procedural skill reuse~\cite{fang2025memp}. Despite these advances, imposing explicit graph structures on natural dialogue introduces inherent trade-offs. Compressing fluid narratives into rigid entity-relation triplets often incurs semantic loss~\cite{chaudhri2022knowledge,zhuang2025linearrag}, while extraction instability allows hallucinated relations to propagate as persistent structural noise~\cite{zhong2021frustratingly,kolluru2020openie6}. The computational overhead of continuous graph maintenance further poses latency challenges for real-time agentic applications~\cite{edge2024local,fang2025lightmem}.

A parallel line of research seeks a middle ground by enabling structured consolidation without rigid graph schemas. HiMem~\cite{zhang2026himem} organizes memory into hierarchical text segments bounded by physical session boundaries, optimizing for compression and retrieval indexing. TiMem~\cite{li2026timem} introduces per-turn reflective thinking chains to deepen single-turn understanding, though at the cost of continuous per-turn overhead. PREMem~\cite{kim2025pre} shifts inference burden to the memory stage by pre-reasoning user preferences before storage, targeting long-term persona consistency. EMem~\cite{zhou2025simple} prioritizes retrieval faithfulness through raw episode preservation, relying on retrieval-driven passive consolidation rather than active synthesis. MemWeaver~\cite{yu2025memweaver} introduces lightweight entity extraction to organize experiences at the session level.
\section{Method}

\begin{figure*}[!t]
    \centering
    \includegraphics[width=1.0\linewidth]{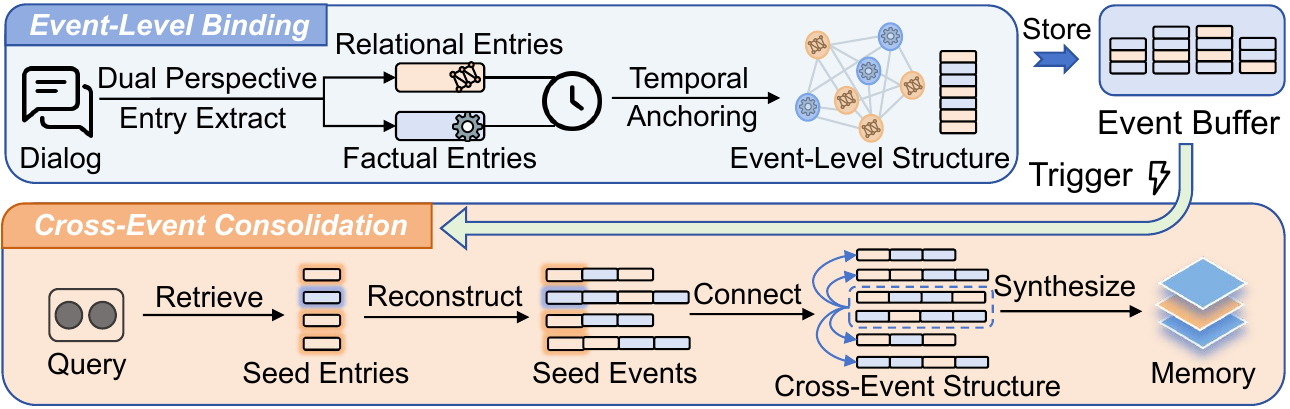}
    \caption{StructMem's hierarchical memory organization. \textbf{Event-Level Binding} constructs event-level structure by extracting dual perspectives and anchoring them temporally. \textbf{Cross-Event Consolidation} constructs cross-event structure through semantic retrieval, event reconstruction, and consolidation synthesis.}
    \label{fig:overview}
\end{figure*}

We propose \textbf{StructMem}, a framework that achieves structure-enriched organization through hierarchical design. The framework operates at two levels: event-level structure (\S\ref{sec:event-level}) preserves relational bindings within individual utterances, while cross-event structure (\S\ref{sec:cross-event}) connects information across temporal boundaries.

\subsection{Event-Level Binding}
\label{sec:event-level}
Event-level binding preserves the connection between factual content and relational context within individual utterances through dual-perspective extraction and temporal anchoring.

\textbf{Dual-Perspective Extraction.} 
For each utterance $m_i$ in the dialogue stream, we extract entries from two complementary perspectives using language model $\mathcal{L}$ with prompts $P_{fact}$ and $P_{rel}$:
\begin{equation}
\Phi_i \cup \Psi_i = \mathcal{L}( P_{fact} \| m_i ) \cup \mathcal{L}( P_{rel} \| m_i ),
\end{equation}
where $\Phi_i = \{c_{i,1}, \ldots, c_{i,j}\}$ contains \textit{factual entries} describing event content, and $\Psi_i = \{r_{i,1}, \ldots, r_{i,k}\}$ contains \textit{relational entries} capturing interpersonal dynamics, causal influences, and temporal dependencies. 
 
By representing both in natural language rather than rigid triplets, we preserve the contextual nuances required for episodic grounding while avoiding entity resolution overhead.
  
\textbf{Temporal Anchoring.}
To preserve the binding between relational and factual information, all entries are anchored to their originating timestamp $\tau_i$, forming an event-level unit:
\begin{equation}
\mathcal{M} \leftarrow  \bigcup_{i=1}^{N} \left\{ \langle x, \mathbf{e}_x, \tau_i \rangle \mid x \in \Phi_i \cup \Psi_i \right\},
\end{equation}
where $\mathbf{e}_x$ denotes the embedding of entry $x$. This temporal coupling enables reconstruction of complete factual-relational events during retrieval.

\begin{table*}[t]
\centering
\resizebox{\linewidth}{!}{%
\begin{tabular}{l|c|cccc|ccc|c|c}
\toprule[1pt]
\multirow{2}{*}{\textbf{Method}} & \multirow{2}{*}{\textbf{Overall $\uparrow$}} & \multicolumn{4}{c|}{\textbf{Performance by Type $\uparrow$}} & \multicolumn{3}{c|}{\textbf{Build Tokens (M) $\downarrow$}} & \multirow{2}{*}{\textbf{Calls $\downarrow$}} & \multirow{2}{*}{\textbf{Time (s) $\downarrow$}} \\
\cmidrule(lr){3-6} \cmidrule(lr){7-9}
& & \textbf{Multi} & \textbf{Open} & \textbf{Single} & \textbf{Temp} & \textbf{In} & \textbf{Out} & \textbf{Sum} & & \\
\midrule
\rowcolor{RAGColor} OpenAI & 71.82 & 69.86 & 53.12 & \underline{84.66} & 45.48 & -- & -- & -- & -- & -- \\
\rowcolor{RAGColor} FullContext &73.83	&68.79	&56.25	&\textbf{86.56}	&50.16 & -- & -- & -- & -- & -- \\
\rowcolor{RAGColor} MiniRAG & 63.51 & 56.74 & 58.33 & 75.74 & 38.94 & 9.022 & 1.081 & 10.103 & \underline{2508} & \textbf{2566} \\
\rowcolor{RAGColor} LightRAG & 68.83 & 66.31 & 50.00 & 77.53 & 53.89 & 10.014 & 1.916 & 11.931 & 13576 & 60469 \\
\midrule
\rowcolor{MemoryColor} LangMem & 58.10 & 62.23 & 47.92 & 71.12 & 23.43 & 9.873 & 1.192 & 11.066 & 5990 & 26281 \\
\rowcolor{MemoryColor} A-Mem &64.16	&56.03	&31.25	&72.06	&60.44 & 9.126 & 2.368 & 11.494 & 11754 & 60607 \\
\rowcolor{MemoryColor} Mem0 &66.88 &67.13 & 51.15	&72.93 &59.19 & 10.958 & 1.239 & 12.196 & 9181 & 30057 \\

\midrule
\rowcolor{StructuralMemoryColor} MemoryOS & 58.25 & 56.74	& 45.83	& 67.06	& 40.19 & \underline{1.889} & \underline{0.939} & \underline{2.868} & 5534 & 24220\\
\rowcolor{StructuralMemoryColor} 
$\text{Mem0}^\text{g}$ & 68.44 &65.71 &47.19 &75.71	&58.13 & 33.512 & 2.313 & 35.825 & 53514 & 115670 \\
\rowcolor{StructuralMemoryColor} Zep & 75.14 & \textbf{74.11} & \textbf{66.04} & 79.79 & 67.71 & -- & -- & -- & -- & -- \\
\rowcolor{StructuralMemoryColor} Memobase & \underline{75.78} & \underline{70.92} & 46.88 & 77.17 & \textbf{85.05} & -- & -- & -- & -- & -- \\
\rowcolor{StructuralMemoryColor} \textbf{StructMem} & \textbf{76.82} & 68.77 & 46.88 & 81.09 & \underline{81.62} & \textbf{1.501} & \textbf{0.436} & \textbf{1.937} & \textbf{1056} & \underline{22854} \\
\bottomrule[1pt]
\end{tabular}
} 
\caption{Performance and resource consumption comparison of memory systems on \texttt{LoCoMo} dataset. $\uparrow$: larger is better; $\downarrow$:  smaller is better.  \textbf{The best results} are marked in bold,  \underline{the second-best results} are underlined. Row colors distinguish method categories: \colorbox{RAGColor}{RAG methods}, \colorbox{MemoryColor}{Flat Memory methods}, and \colorbox{StructuralMemoryColor}{Structural Memory methods}. OpenAI and FullContext have no construction cost; Zep and Memobase do not expose construction details.}
\label{tab:locomo}
\end{table*}

\subsection{Cross-Event Consolidation}
\label{sec:cross-event} 
Cross-event consolidation connects information across temporal by periodically synthesizing semantically related events. We trigger synthesis when accumulated events exceed a time threshold.

\textbf{Semantic Event Connections.} 
We buffer unconsolidated entries since the last consolidation. The buffered entries are temporally ordered:
\begin{equation}
\mathcal{C}_{buf} = \text{Sort}_{\tau} \{ x \in \mathcal{M}_{buffer} \},
\end{equation}
where $\mathcal{M}_{buffer}$ denotes the buffered entries. We encode the buffered context into an aggregated query by concatenating all buffered entry texts and encoding them with an embedding model. We then rank all historical entries by cosine similarity to this query and retrieve the top-$K$ most semantically similar entries as seeds, denoted as $\mathcal{S}_k$.

For each seed entry $x^* \in \mathcal{S}_k$, we reconstruct its complete event context by retrieving all entries sharing the same timestamp:
\begin{equation}
{E}_{\tau}(x^*) = \{ x' \in \mathcal{M} \mid \tau(x') = \tau(x^*) \}.
\end{equation}
These reconstructed events, together with the buffered events, form the cross-event structure grounded in semantic relevance.
\begin{equation}
\mathcal{C}_{cross} = \mathcal{C}_{buf} \cup \bigcup_{x^* \in \mathcal{S}_k} {E}_{\tau}(x^*).
\end{equation}
\textbf{Memory Consolidation through Synthesis.} 
Unlike conventional summarization that performs lossy compression on sequential text, our consolidation mechanism operates on semantically-reconstructed event clusters. It explicitly synthesizes cross-event relational hypotheses, forming a complementary abstraction layer that enables multi-hop reasoning while faithfully preserving the fidelity of raw episodic memory.
\begin{equation}
\mathcal{M} \leftarrow \mathcal{C}_{cons} = \mathcal{L}( P_{cons} \| \mathcal{C}_{cross} ).
\end{equation}

\section{Experiments}
\subsection{Experimental Setup}
We first describe the dataset and evaluation metrics, followed by the baseline systems used for comparison. To ensure reproducibility, complete set of prompt templates and implementation details used for memory construction, question answering, and evaluation is provided in Appendix~\ref{appendix:prompts}.
\paragraph{Dataset and Metrics.} We evaluate on the \texttt{LoCoMo} benchmark~\cite{maharana2024evaluating} (see Appendix~\ref{dataset} for detailed statistics). Effectiveness is measured using LLM-as-a-judge evaluation; efficiency is measured by token usage, API calls, and runtime during memory construction.

\paragraph{Baselines.} We compare StructMem against RAG-based systems (OpenAI, FullContext, MiniRAG, LightRAG), flat memory methods (LangMem, A-Mem, Mem0), and structural memory methods (MemoryOS, Mem0$^\text{g}$, Zep, Memobase). All methods use gpt-4o-mini as the backbone and text-embedding-3-small for embeddings. Detailed retrieval and configuration parameters for all baselines are provided in Appendix~\ref{appendix:baseline}.

\subsection{Overall Performance}

Table~\ref{tab:locomo} shows StructMem achieves state-of-the-art overall performance on \texttt{LoCoMo}, with substantial gains in multi-domain and temporal reasoning where cross-event connections are critical for understanding causal relationships across dialogue sessions. Beyond effectiveness, StructMem demonstrates exceptional efficiency: compared to existing memory systems, it reduces token consumption and requires significantly fewer API calls, as our progressive structural organization avoids the expensive post-hoc graph construction. These results hold consistently across multiple judge models, as verified in Appendix~\ref{appendix:robustness}.

\subsection{Analysis}
We analyze StructMem from two complementary perspectives: a paradigm-level comparison that evaluates effectiveness and efficiency across all three memory paradigms, and an internal analysis that examines the mechanism underlying StructMem's reasoning gains.
\begin{figure*}[!t]
    \centering
    \subfigure[Token consumption over dialogue turns] { \label{fig:analysis_a} \includegraphics[width=0.49\linewidth]{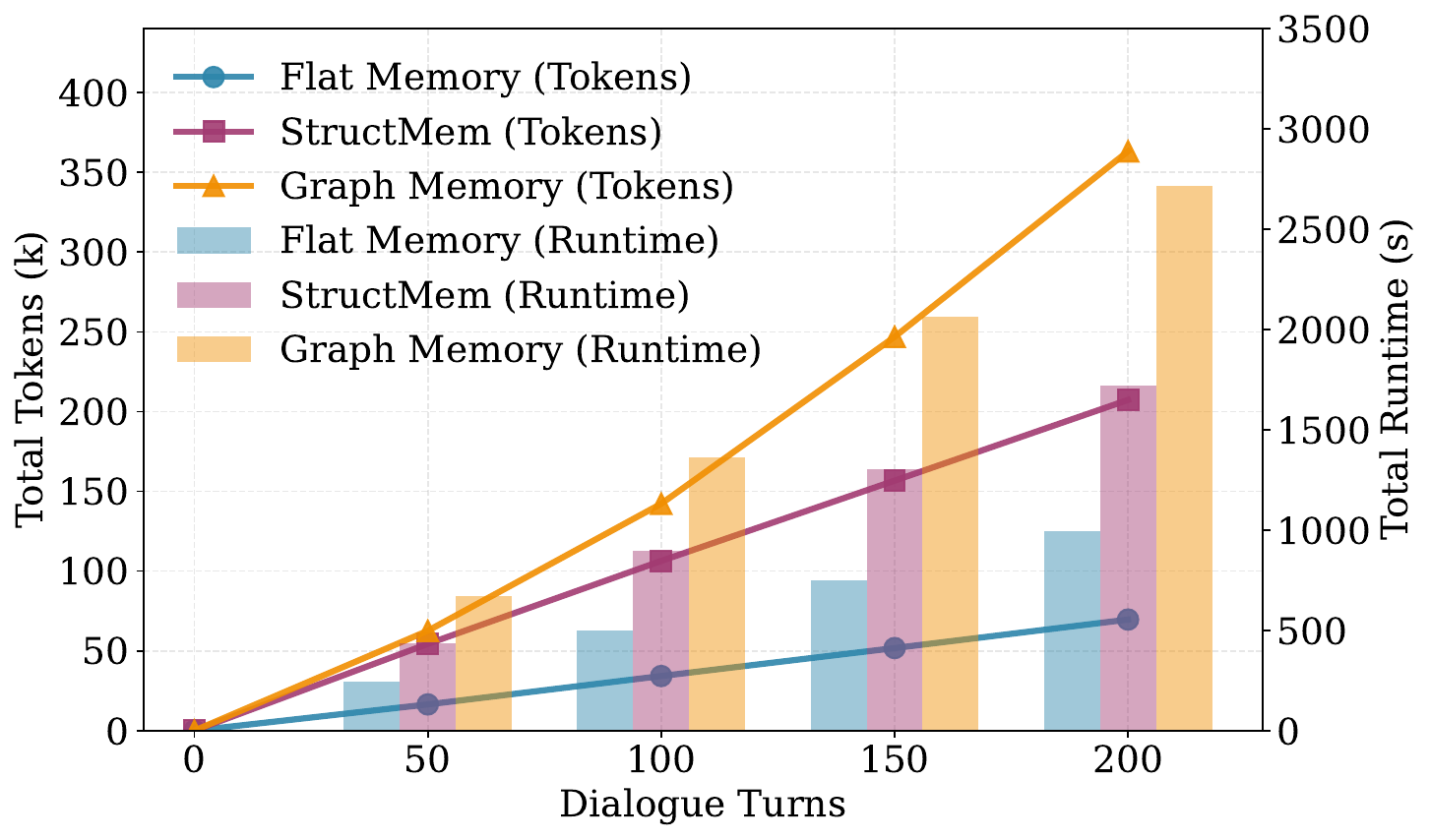}}
    \subfigure[Component-wise token consumption] { \label{fig:analysis_b} \includegraphics[width=0.49\linewidth]{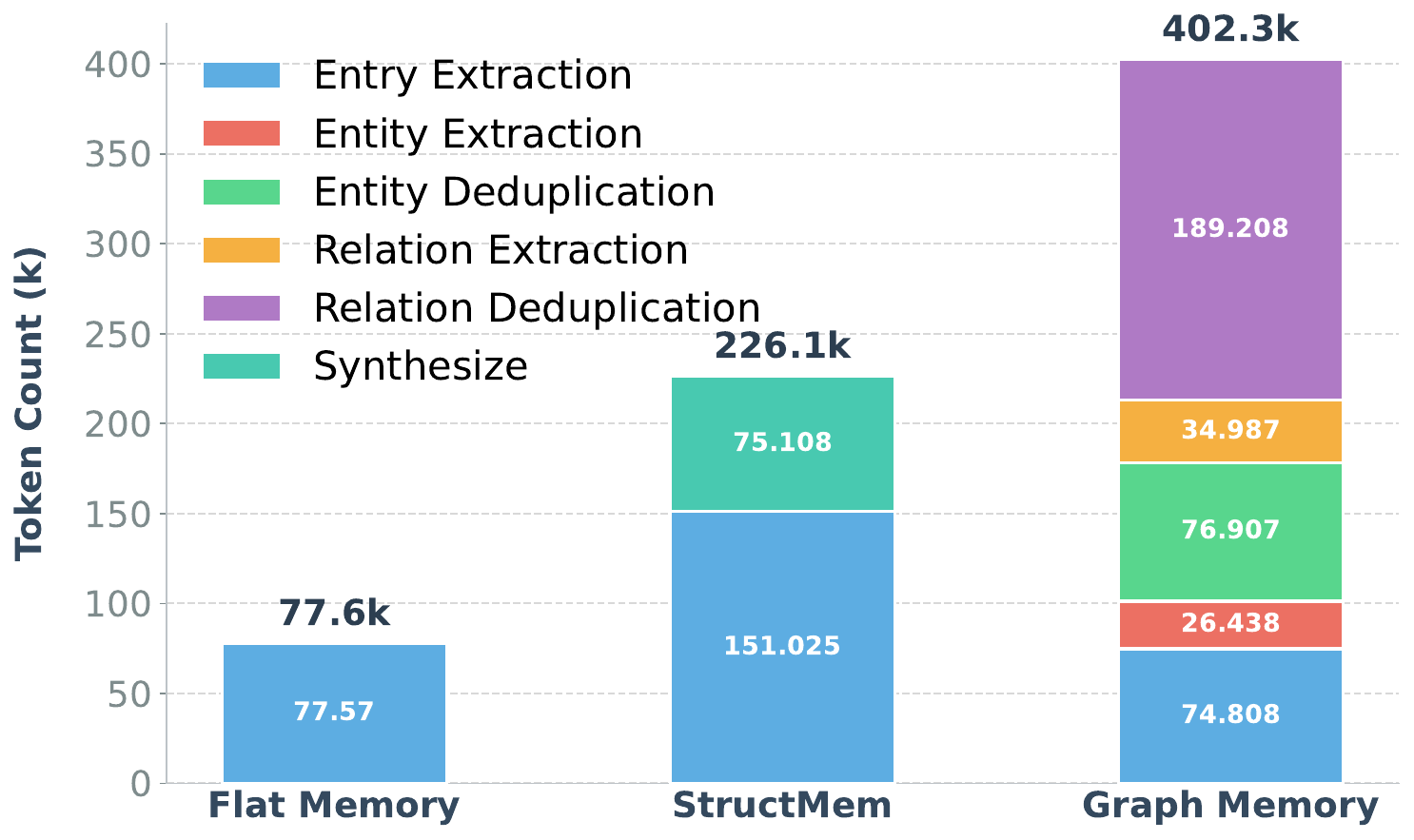}}
    \subfigure[Effect of the number of retrieved entries] { \label{fig:analysis_c} \includegraphics[width=0.49\linewidth]{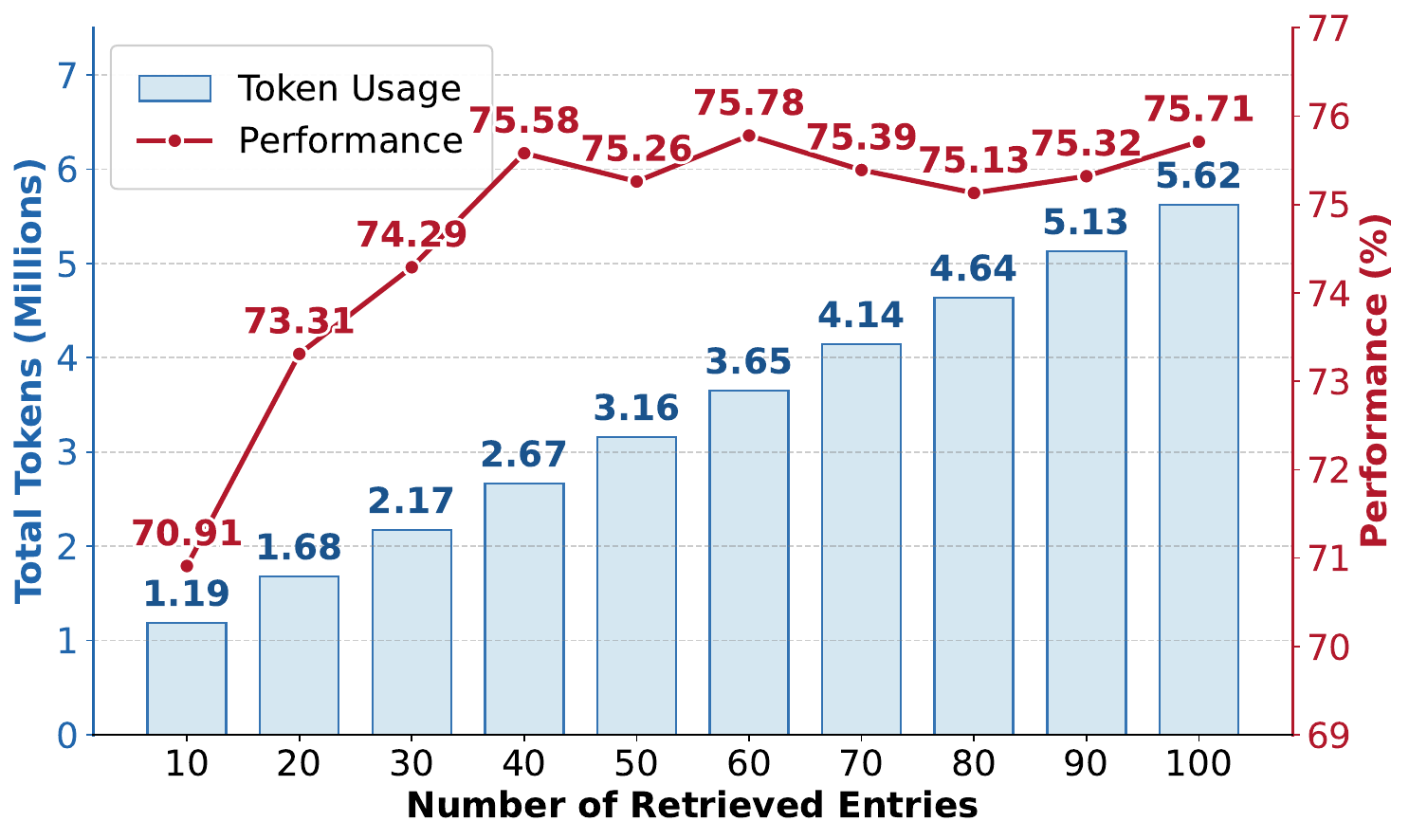}}    
    \subfigure[Effect of the number of semantic retrieval seeds $K$] { 
    \label{fig:analysis_d} \includegraphics[width=0.49\linewidth]{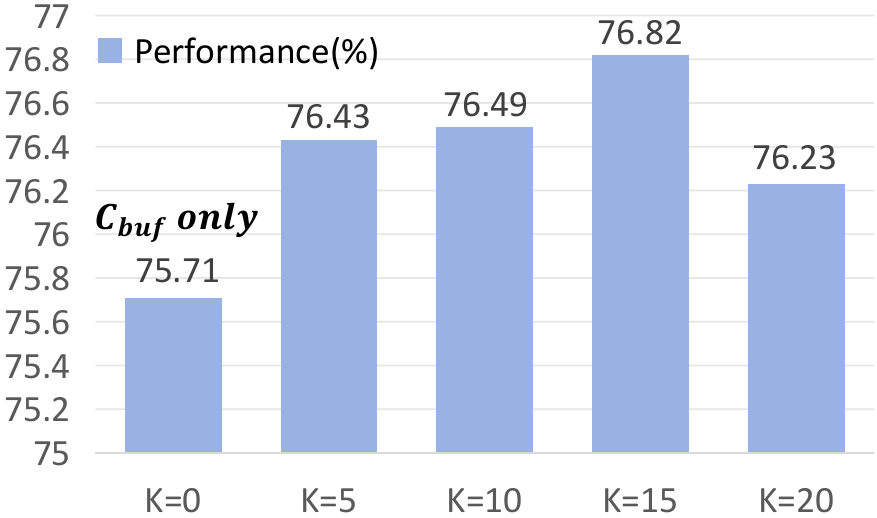}}
    \caption{Analysis of efficiency across memory paradigms and internal mechanisms of StructMem.}
    \label{fig:analysis}
\end{figure*}
\paragraph{Paradigm Comparison.} 

\begin{table}[!htbp]
\centering
\small 
\begin{tabular}{l|cccc}
\toprule
\textbf{Method} & \textbf{Multi} & \textbf{Open} & \textbf{Single} & \textbf{Temp}\\
\midrule
Flat Memory & 66.31 & 46.88 & 78.83 & 78.50 \\
Graph Memory & 66.67 & 48.96 & 80.50 & 76.64 \\  
\midrule
w/o Cross-Event & 66.31 & 46.88 & 80.86 & 79.44 \\
StructMem & 68.77 & 46.88 & 81.09 & 81.62 \\
\bottomrule
\end{tabular}
\caption{Paradigm comparison and ablation study on \texttt{LoCoMo} dataset.}
\label{tab:ablation}
\end{table}

To validate the effectiveness of each paradigm, we conduct studies in Table~\ref{tab:ablation}. Starting from Flat Memory as the baseline, Graph Memory achieves improvements on single-session and open-domain tasks, though it decreases on temporal reasoning. In contrast, our approach demonstrates consistent improvements across all task types. Event-level structure improves performance in temporal reasoning and single-session. Cross-event structure yields further gains by capturing cross-temporal causal relationships.

To examine computational efficiency, we analyze token usage and runtime on the first conversation of \texttt{LoCoMo}. Figure~\ref{fig:analysis_a} shows that Graph Memory incurs significantly higher token usage and runtime as dialogue progresses. Figure~\ref{fig:analysis_b} reveals the source: graph construction requires four cascading LLM operations per event, with deduplication overhead growing quadratically. In contrast, StructMem achieves efficiency through buffered consolidation: by exploiting temporal locality, in which semantically related events naturally cluster within short time windows, the system accumulates events and processes them in batch during periodic synthesis. This effectively reduces cross-event organization from per-event operations to periodic batch processing, substantially cutting both API calls and token consumption.

\paragraph{StructMem Internal Mechanisms.} 
\label{Internal_Mechanisms}
We analyze whether hierarchical organization provides genuine reasoning gains beyond retrieval scaling. 

Figure~\ref{fig:analysis_c} reveals that flat retrieval performance peaks at 60 entries and plateaus thereafter, indicating that simply retrieving more atomic entries cannot improve effectiveness, as the bottleneck is knowledge reasoning rather than coverage. Cross-event consolidation addresses this by synthesizing semantically related events into higher-level relational hypotheses, creating information that does not exist in any individual memory entry.

Figure~\ref{fig:analysis_d} confirms this: without event connections ($K=0$), performance matches the flat retrieval plateau, but introducing cross-event synthesis yields substantial gains, demonstrating that hierarchical consolidation reconstructs causal relationships across temporal boundaries and enables fundamentally new reasoning capabilities. Fidelity analyses in Appendix~\ref{appendix:fidelity} further confirm that these synthesized connections are well-grounded, with minimal spurious associations.
\section{Conclusion}
We propose StructMem, which achieves structure-enriched organization through hierarchical design: preserving event-level bindings and enabling cross-event consolidation, StructMem preserves temporal and relational structures without the computational overhead of continuous graph maintenance. Experiments on \texttt{LoCoMo} demonstrate that StructMem achieves better performance with strong results in multi-hop and temporal reasoning, while substantially reducing token consumption, API calls, and runtime compared to prior memory systems.

\section*{Limitations}
Despite its strong performance, StructMem has several limitations. The quality of dual-perspective extraction is highly dependent on instruction prompts, where suboptimal design may result in incomplete or inaccurate relational information capture. Future research could investigate automated prompt optimization to improve robustness across various dialogue contexts. Additionally, the framework primarily addresses memory expansion and synthesis but currently lacks an explicit mechanism for conflict resolution and memory updating. As user facts or preferences may evolve over long horizons, the absence of a revision process could lead to inconsistencies between historical summaries and new information. Future iterations should incorporate memory decay or updating strategies to ensure the hierarchical organization accurately reflects the most current state of the interaction.

\section*{Acknowledgements}
We would like to express sincere gratitude to the  reviewers for their thoughtful and constructive feedback. This work was supported by the National Natural Science Foundation of China (No. 62576307), Yongjiang Talent Introduction Programme (2021A-156-G), and Information Technology Center and State Key Lab of CAD\&CG, Zhejiang University. This work was supported by Ant Group and Zhejiang University - Ant Group Joint Laboratory of Knowledge Graph.

\bibliography{sample-base}

\appendix
\section{Appendix}

\subsection{License} \label{license}
This work uses the LoCoMo benchmark dataset, which is publicly available for academic research purposes. We follow all usage terms specified by the dataset authors.

\subsection{Dataset} \label{dataset}
We evaluate on the \textbf{LoCoMo} benchmark~\cite{maharana2024evaluating}, which contains 10 long-term conversations with an average of 588 turns and 16,618 tokens per conversation. We focus on the question answering task, utilizing four reasoning types from the benchmark. Table~\ref{tab:locomo_stats} shows the statistics of questions used in our evaluation. Model performance is evaluated using LLM-as-a-judge.
\begin{table}[t]
\centering
\begin{tabular}{l|c}
\toprule
\textbf{Reasoning Type} & \textbf{\# Questions} \\
\midrule
Single-hop & 841 \\
Multi-hop & 282 \\
Temporal & 321 \\
Open-domain & 96 \\
\bottomrule
\end{tabular}
\caption{Statistics of LoCoMo questions used.}
\label{tab:locomo_stats}
\end{table}
\definecolor{JudgeGPT}{RGB}{235, 245, 255}  
\definecolor{JudgeQwen}{RGB}{255, 245, 235} 
\definecolor{JudgeDS}{RGB}{245, 255, 235}   

\begin{table*}[t]
\centering
\caption{Robustness check of memory systems across different LLM judges on the \texttt{LoCoMo} dataset. The table is categorized by three judge models: \texttt{gpt-4o-mini}, \texttt{Qwen2.5-32B-Instruct}, and \texttt{DeepSeek-V3.2}. \textbf{Bold} and \underline{underline} denote the best and second-best results within each judge block, respectively. $\uparrow$: larger is better.}
\label{tab:robustness_blocks}
\resizebox{\linewidth}{!}{%
\begin{tabular}{llccccc}
\toprule
\textbf{Judge Model} & \textbf{Method} & \textbf{Overall $\uparrow$} & \textbf{Single Hop} & \textbf{Multi Hop} & \textbf{Temporal} & \textbf{Open Domain} \\
\midrule
\rowcolor{JudgeGPT} & FullContext & 73.83 & \textbf{86.56} & 68.79 & 50.16 & \underline{56.25} \\
\rowcolor{JudgeGPT} & A-MEM & 64.16 & 72.06 & 56.03 & 60.44 & 31.25 \\
\rowcolor{JudgeGPT} & MemoryOS & 58.25 & 67.06 & 56.74 & 40.19 & 45.83 \\
\rowcolor{JudgeGPT} & Memobase & \underline{75.78} & 77.17 & \underline{70.92} & \textbf{85.05} & 46.88 \\
\rowcolor{JudgeGPT} & Zep & 75.14 & 79.79 & \textbf{74.11} & 67.71 & \textbf{66.04} \\
\rowcolor{JudgeGPT} \multirow{-6}{*}{\texttt{gpt-4o-mini}} & \textbf{StructMem} & \textbf{76.82} & \underline{81.09} & 68.77 & \underline{81.62} & 46.88 \\
\midrule
\rowcolor{JudgeQwen} & FullContext & 71.17 & \textbf{83.83} & 67.02 & 48.29 & \underline{48.96} \\
\rowcolor{JudgeQwen} & A-MEM & 60.26 & 68.85 & 53.19 & 52.65 & 31.25 \\
\rowcolor{JudgeQwen} & MemoryOS & 60.32 & 69.80 & 62.41 & 39.88 & 39.58 \\
\rowcolor{JudgeQwen} & Memobase & \underline{76.36} & 78.00 & \underline{71.28} & \textbf{85.05} & 47.92 \\
\rowcolor{JudgeQwen} & Zep & 75.52 & 79.19 & \textbf{75.18} & 70.40 & \textbf{61.46} \\
\rowcolor{JudgeQwen} \multirow{-6}{*}{\texttt{Qwen2.5-32B-Instruct}} & \textbf{StructMem} & \textbf{77.01} & \underline{82.16} & \underline{69.86} & \underline{78.19} & \underline{48.96} \\
\midrule
\rowcolor{JudgeDS} & FullContext & 70.97 & \textbf{85.49} & 67.02 & 41.43 & \underline{54.17} \\
\rowcolor{JudgeDS} & A-MEM & 63.90 & 74.55 & 56.38 & 47.35 & 47.92 \\
\rowcolor{JudgeDS} & MemoryOS & 61.56 & 72.53 & 62.06 & 35.83 & 50.00 \\
\rowcolor{JudgeDS} & Memobase & \underline{79.29} & 80.86 & \textbf{77.66} & \textbf{85.67} & 48.96 \\
\rowcolor{JudgeDS} & Zep & 75.45 & 80.98 & \underline{75.89} & 64.80 & \textbf{61.46} \\
\rowcolor{JudgeDS} \multirow{-6}{*}{\texttt{DeepSeek-V3.2}} & \textbf{StructMem} & \textbf{79.35} & \underline{85.14} & 73.05 & \underline{77.57} & 53.12 \\
\bottomrule
\end{tabular}
}
\end{table*}
\begin{table}[ht]
\centering
\caption{Inter-judge agreement and correlation across different judge model pairs. \texttt{GPT}, \texttt{Qwen}, and \texttt{DS} denote \texttt{gpt-4o-mini}, \texttt{qwen2.5-32b-instruct}, and \texttt{DeepSeek-V3.2}, respectively.}
\label{tab:inter_judge_stats}
\resizebox{\linewidth}{!}{%
\begin{tabular}{lccc}
\toprule
\textbf{Judge Pair} & \textbf{Cohen's $\kappa$} & \textbf{Pearson $r$} & \textbf{$p$-value} \\
\midrule
\texttt{Qwen} vs. \texttt{DS}   & 0.8395 & 0.8438 & $< 10^{-300}$ \\
\texttt{Qwen} vs. \texttt{GPT} & 0.8326 & 0.8362 & $< 10^{-300}$ \\
\texttt{DS} vs. \texttt{GPT}   & 0.8184 & 0.8234 & $< 10^{-300}$ \\
\midrule
\textbf{Overall (Fleiss' $\kappa$)} & \textbf{0.8341} & — & — \\
\bottomrule
\end{tabular}
}
\end{table}

\subsection{Implementation Details}
We provide key implementation details for StructMem to facilitate reproducibility. 
\textbf{Memory construction:} We set the time window threshold to 1 hour for triggering consolidation. For cross-event consolidation, we retrieve top-15 semantically similar seed entries from historical memory. \textbf{Question answering:} During inference, we retrieve 60 entries and 5 synthesis from memory to provide context for answer generation. 

\subsection{Baseline Configurations} \label{appendix:baseline}

To ensure empirical rigor and reproducibility, we provide the detailed retrieval and architectural configurations for all evaluated systems:

\textbf{FullContext} utilizes the entire raw dialogue history fed into the prompt in reverse chronological order via a full-scan with $k=-1$. \textbf{OpenAI} processes all conversation turns concatenated as a flat, unordered text sequence directly without a retrieval step.

\textbf{MiniRAG} and \textbf{LightRAG} retrieve the top-20 relevant entries per question to provide factual context. Similarly, \textbf{A-MEM} and \textbf{LangMem} employ a global search mechanism to retrieve the top-40 most relevant memory entries for each query.

\textbf{MemoryOS} implements a three-tier hierarchical system featuring exhaustive recall of all Short-Term Memory (STM) pages, a two-stage selection for Mid-Term Memory (MTM) comprising the top-5 segments and top-10 dialogue pages, and the extraction of the top-10 relevant entries from Long-term Personal Memory (LPM).

For API-based systems including \textbf{Mem0}, \textbf{Mem0$^\text{g}$}, \textbf{Zep}, and \textbf{Memobase}, the top-10 relevant memories per speaker are retrieved for response generation.

\subsection{Robustness of Evaluation}\label{appendix:robustness}

We validate the reliability of our LLM-as-a-judge protocol by conducting extensive cross-model evaluations and statistical analyses. 

Table~\ref{tab:robustness_blocks} summarizes the performance of memory systems across three distinct judge model families: gpt-4o-mini, Qwen2.5-32b-Instruct, and DeepSeek-V3.2. We further calculate the inter-judge agreement and correlation across all judge pairs, as detailed in Table~\ref{tab:inter_judge_stats}. The Fleiss' $\kappa$ among different models reaches \textbf{0.8341}, reflecting a near-perfect agreement that substantially exceeds the commonly accepted reliability threshold of 0.8. This high level of consensus, combined with significant Pearson correlation coefficients ($r > 0.81$, $p < 10^{-300}$), confirms that the automated evaluation protocol provides a stable and objective assessment of semantic response quality.

\subsection{Fidelity and Hallucination Study}\label{appendix:fidelity}

We conducted a systematic study to ensure that the induced structures are grounded in the source dialogue. 

\begin{table}[ht]
\centering
\caption{Hallucination rates in the event-level extraction stage across 10 conversations.}
\label{tab:extraction_hallucination}
\resizebox{\linewidth}{!}{%
\begin{tabular}{lcccc}
\toprule
\textbf{Conversation} & \textbf{DS} & \textbf{Qwen} & \textbf{GPT} & \textbf{Mean} \\
\midrule
conv-26 & 2.07\% & 0.52\% & 0.78\% & 1.12\% \\
conv-30 & 1.81\% & 0.60\% & 4.83\% & 2.41\% \\
conv-41 & 2.04\% & 1.88\% & 2.35\% & 2.09\% \\
conv-42 & 3.16\% & 1.97\% & 3.94\% & 3.02\% \\
conv-43 & 2.94\% & 1.63\% & 3.10\% & 2.56\% \\
conv-44 & 1.68\% & 0.92\% & 1.68\% & 1.43\% \\
conv-47 & 5.28\% & 2.44\% & 3.05\% & 3.59\% \\
conv-48 & 2.71\% & 1.45\% & 1.08\% & 1.75\% \\
conv-49 & 3.25\% & 1.16\% & 1.62\% & 2.01\% \\
conv-50 & 3.55\% & 2.84\% & 4.26\% & 3.55\% \\
\midrule
\textbf{Overall} & \textbf{2.84\%} & \textbf{1.61\%} & \textbf{2.63\%} & \textbf{2.36\%} \\
\bottomrule
\end{tabular}
}
\end{table}

\paragraph{Event-Level Extraction Fidelity.} We first evaluated whether the atomic memory entries  accurately reflect the original utterances. We employed three independent judge models (gpt-4o-mini, Qwen2.5-32B-Instruct, and DeepSeek-V3.2) to identify hallucinated entries across conversations. 

Specifically, for each extracted memory entry, the judges are provided with the corresponding source dialogue segment and tasked with determining if any factual or relational information is fabricated or unsubstantiated by the original text. The full prompt templates are provided in Figure~\ref{fig:Extraction_Fidelity}.

As detailed in Table~\ref{tab:extraction_hallucination}, the mean hallucination rate is only 2.36\%, confirming that the \textbf{Dual-Perspective Extraction} of our hierarchical memory is highly faithful to the source context. 

\begin{table*}[t]
\centering
\caption{Detailed cross-event link quality comparison across conversations. \textbf{S}, \textbf{T}, and \textbf{R} denote the number of \textbf{S}purious links, \textbf{T}otal links, and the error \textbf{R}ate (\%), respectively. \texttt{GPT}, \texttt{Qwen}, and \texttt{DS} represent \texttt{gpt-4o-mini}, \texttt{Qwen2.5-32B-Instruct}, and \texttt{DeepSeek-V3.2}. \textit{Constrained} is the default setting for StructMem.}
\label{tab:consolidation_fidelity}
\resizebox{\linewidth}{!}{%
\begin{tabular}{ll|ccc|ccc|ccc}
\toprule
\multirow{2}{*}{\textbf{Conversation}} & \multirow{2}{*}{\textbf{Config}} & \multicolumn{3}{c|}{\textbf{Judge: GPT}} & \multicolumn{3}{c|}{\textbf{Judge: Qwen}} & \multicolumn{3}{c}{\textbf{Judge: DS}} \\
\cmidrule(lr){3-5} \cmidrule(lr){6-8} \cmidrule(lr){9-11}
& & \textbf{S} & \textbf{T} & \textbf{R (\%)} & \textbf{S} & \textbf{T} & \textbf{R (\%)} & \textbf{S} & \textbf{T} & \textbf{R (\%)} \\
\midrule
\multirow{2}{*}{conv-26} & Constrained & 0 & 83 & 0.00 & 3 & 70 & 4.29 & 1 & 12 & 8.33 \\
& Unconstrained & 4 & 72 & 5.56 & 23 & 101 & 22.77 & 10 & 58 & 17.24 \\
\midrule
\multirow{2}{*}{conv-30} & Constrained & 0 & 73 & 0.00 & 0 & 59 & 0.00 & 1 & 23 & 4.35 \\
& Unconstrained & 4 & 84 & 4.76 & 6 & 79 & 7.59 & 2 & 44 & 4.55 \\
\midrule
\multirow{2}{*}{conv-41} & Constrained & 4 & 108 & 3.70 & 1 & 131 & 0.76 & 1 & 31 & 3.23 \\
& Unconstrained & 13 & 104 & 12.50 & 20 & 115 & 17.39 & 6 & 74 & 8.11 \\
\midrule
\multirow{2}{*}{conv-42} & Constrained & 0 & 95 & 0.00 & 5 & 93 & 5.38 & 1 & 47 & 2.13 \\
& Unconstrained & 8 & 100 & 8.00 & 20 & 106 & 18.87 & 6 & 65 & 9.23 \\
\midrule
\multirow{2}{*}{conv-43} & Constrained & 0 & 104 & 0.00 & 5 & 130 & 3.85 & 6 & 40 & 15.00 \\
& Unconstrained & 4 & 109 & 3.67 & 11 & 114 & 9.65 & 9 & 74 & 12.16 \\
\midrule
\multirow{2}{*}{conv-44} & Constrained & 0 & 110 & 0.00 & 9 & 91 & 9.89 & 2 & 39 & 5.13 \\
& Unconstrained & 6 & 104 & 5.77 & 30 & 135 & 22.22 & 16 & 88 & 18.18 \\
\midrule
\multirow{2}{*}{conv-47} & Constrained & 1 & 107 & 0.93 & 4 & 90 & 4.44 & 0 & 33 & 0.00 \\
& Unconstrained & 11 & 103 & 10.68 & 32 & 108 & 29.63 & 16 & 69 & 23.19 \\
\midrule
\multirow{2}{*}{conv-48} & Constrained & 1 & 102 & 0.98 & 2 & 101 & 1.98 & 0 & 36 & 0.00 \\
& Unconstrained & 5 & 100 & 5.00 & 27 & 107 & 25.23 & 11 & 62 & 17.74 \\
\midrule
\multirow{2}{*}{conv-49} & Constrained & 0 & 94 & 0.00 & 2 & 90 & 2.22 & 0 & 52 & 0.00 \\
& Unconstrained & 7 & 81 & 8.64 & 14 & 86 & 16.28 & 10 & 61 & 16.39 \\
\midrule
\multirow{2}{*}{conv-50} & Constrained & 0 & 106 & 0.00 & 2 & 113 & 1.77 & 1 & 45 & 2.22 \\
& Unconstrained & 10 & 109 & 9.17 & 30 & 114 & 26.32 & 18 & 92 & 19.57 \\
\midrule
\midrule
\multirow{2}{*}{\textbf{Overall}} & \textbf{Constrained} & \textbf{6} & \textbf{982} & \textbf{0.61\%} & \textbf{33} & \textbf{968} & \textbf{3.41\%} & \textbf{13} & \textbf{358} & \textbf{3.63\%} \\
& \textbf{Unconstrained} & \textbf{72} & \textbf{966} & \textbf{7.45\%} & \textbf{213} & \textbf{1065} & \textbf{20.00\%} & \textbf{104} & \textbf{687} & \textbf{15.14\%} \\
\bottomrule
\end{tabular}
}
\end{table*}

\paragraph{Cross-Event Consolidation Fidelity.} The most critical verification involves the synthesis of cross-event links. To isolate and audit these links, we employed three independent judge models (gpt-4o-mini, Qwen2.5-32B-Instruct, and DeepSeek-V3.2) to identify hallucinated links across conversations. 

For each consolidation step, the judge is provided with: (1) \textit{Buffer Text} (current events), (2) \textit{Supplementary Text} (retrieved history), and (3) two summaries, including \textbf{Summary A} (Baseline, $k=0$, consolidates only buffer events) and \textbf{Summary B} (Test, $k=15$, establishes cross-event links). The judge identifies cross-event links present in Summary B that are absent from Summary A, then classifies each cross-event link to judge if the link is spurious. The full prompt templates are provided in Figure~\ref{fig:Consolidation_Fidelity_1} and Figure~\ref{fig:Consolidation_Fidelity_2}.

To evaluate the specific impact of our grounding anchors, we conduct a sensitivity analysis by comparing our default \textit{Constrained} prompt against an \textit{Unconstrained} variant. As illustrated in Figure~\ref{fig:synthesis_unconstrained_prompt}, the \textit{Unconstrained} version is created by removing explicit requirements for timestamp citations and concrete dependency focus (highlighted in gray).

The results in Table~\ref{tab:consolidation_fidelity} demonstrate that removing these grounding constraints leads to a dramatic surge in hallucination rates across all judge models. This trend underscores that the high fidelity of StructMem's hierarchical organization is directly tied to our constrained synthesis mechanism, confirming that the \textbf{Memory Consolidation} of our hierarchical memory is highly faithful to the source context.

\subsection{Prompt Templates}\label{appendix:prompts}
We present the prompt templates used for memory construction, question answering, and evaluation in StructMem.

\begin{figure*}[t]
    \centering
    \includegraphics[width=0.9\textwidth]{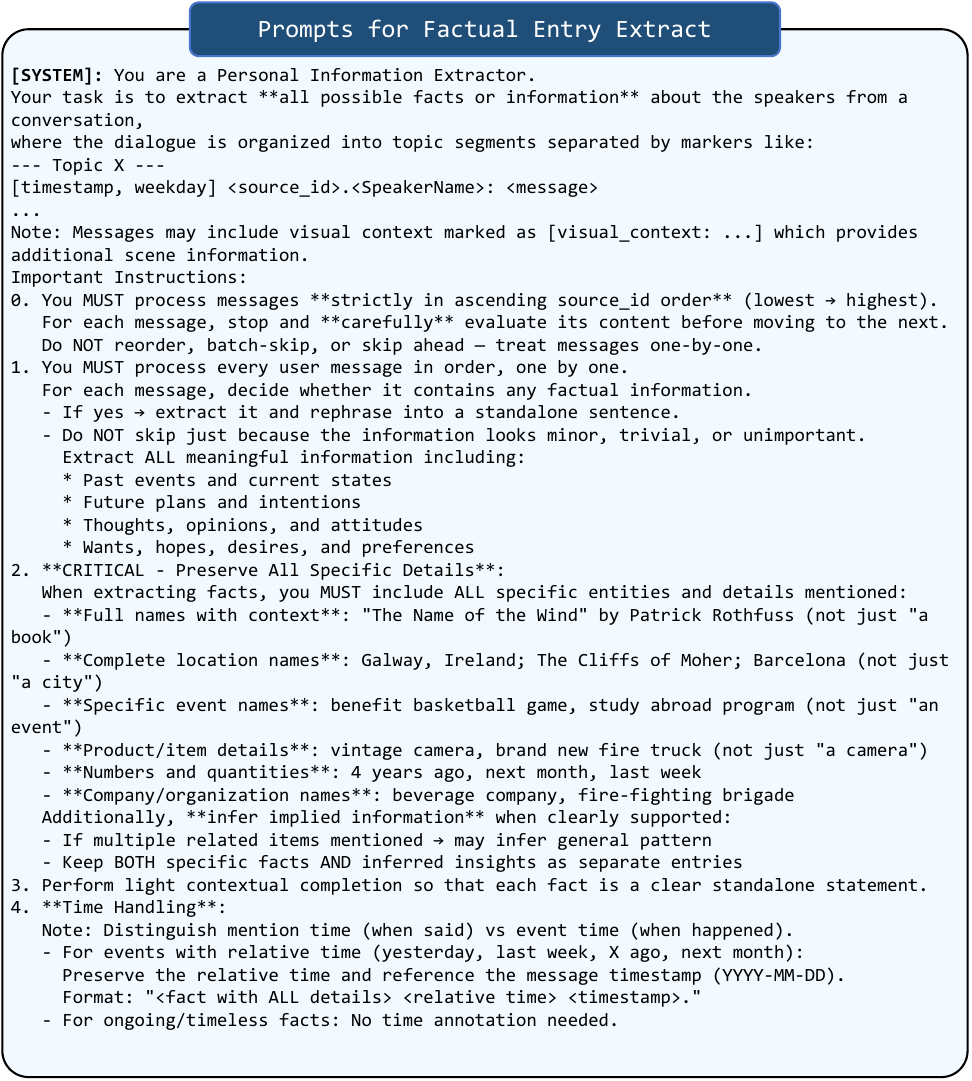}
    \caption{Factual entry extraction prompt (Part 1).}
    \label{fig:fact_1}
\end{figure*}

\begin{figure*}[t]
    \centering
    \includegraphics[width=0.9\textwidth]{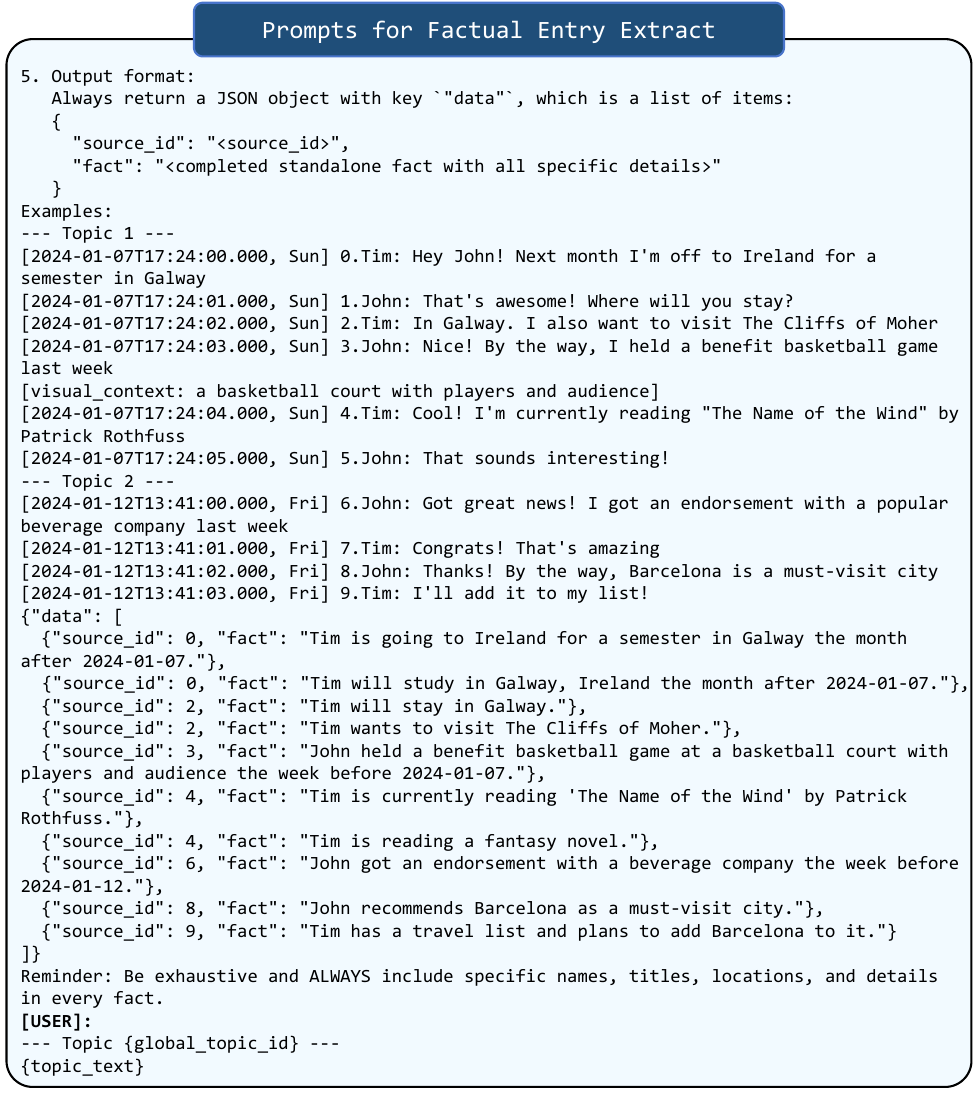}
    \caption{Factual entry extraction prompt (Part 2).}
    \label{fig:fact_2}
\end{figure*}

\begin{figure*}[t]
    \centering
    \includegraphics[width=0.9\textwidth]{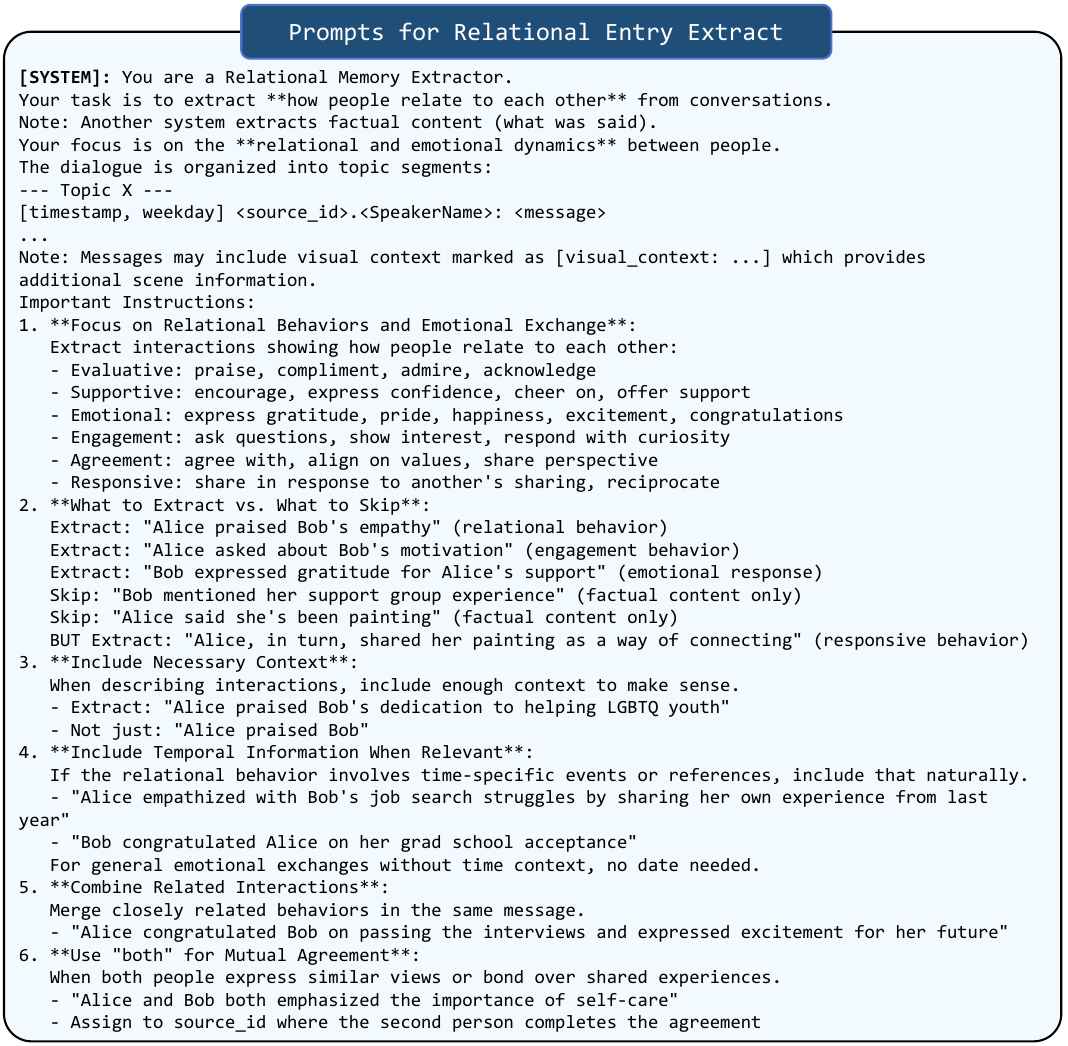}
    \caption{Relational entry extraction prompt (Part 1).}
    \label{fig:relation_1}
\end{figure*}

\begin{figure*}[t]
    \centering
    \includegraphics[width=0.9\textwidth]{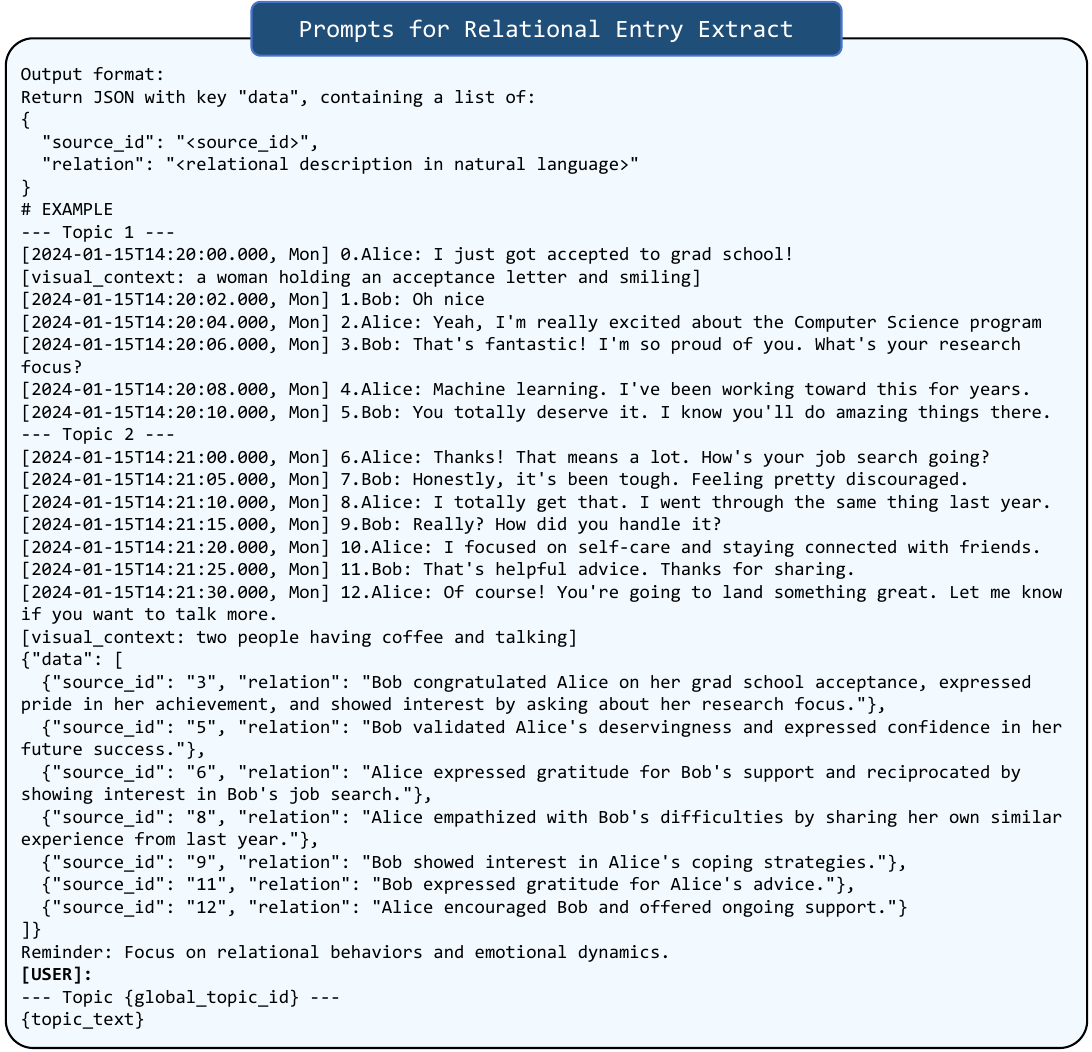}
    \caption{Relational entry extraction prompt (Part 2).}
    \label{fig:relation_2}
\end{figure*}

\begin{figure*}[t]
    \centering
    \includegraphics[width=0.9\textwidth]{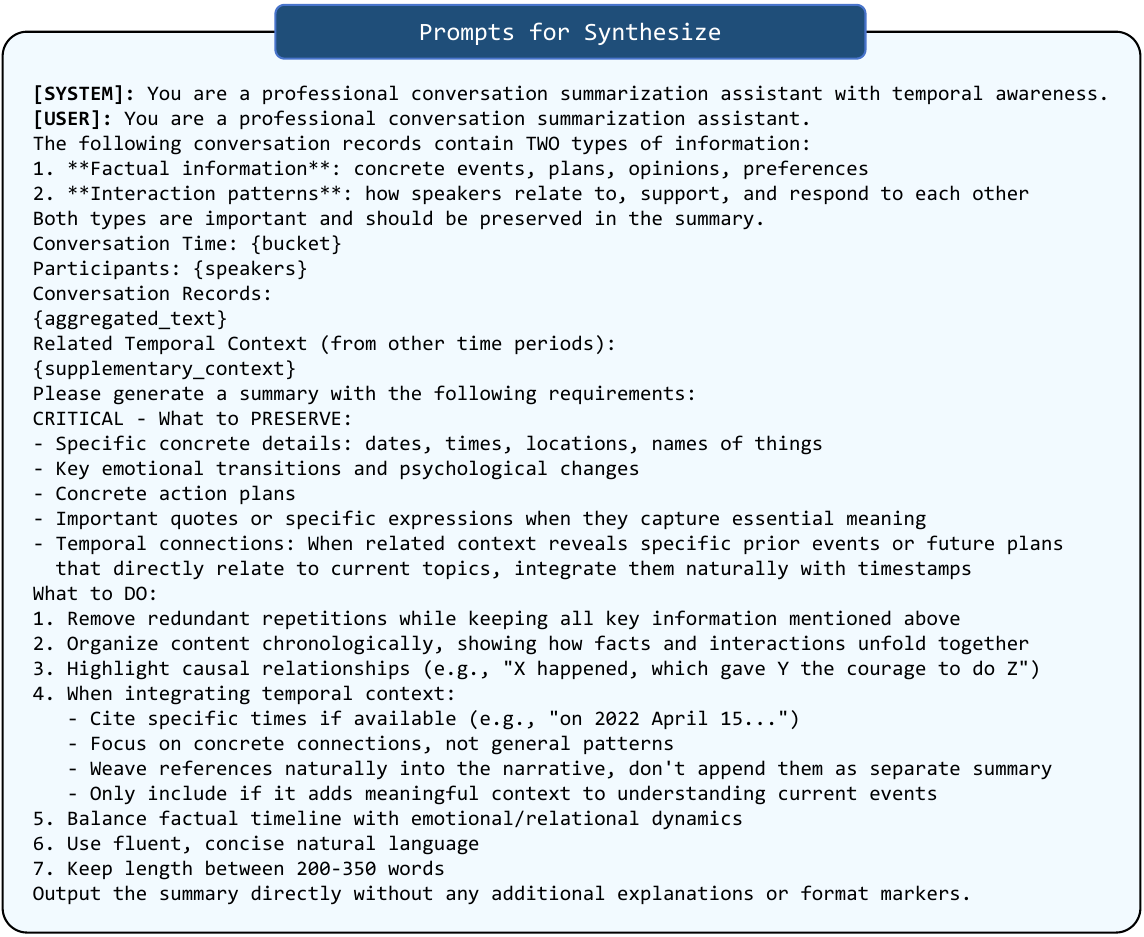}
    \caption{Narrative synthesis prompt for Synthesis.}
    \label{fig:synthesis_prompt}
\end{figure*}

\begin{figure*}[t]
    \centering
    \includegraphics[width=0.9\textwidth]{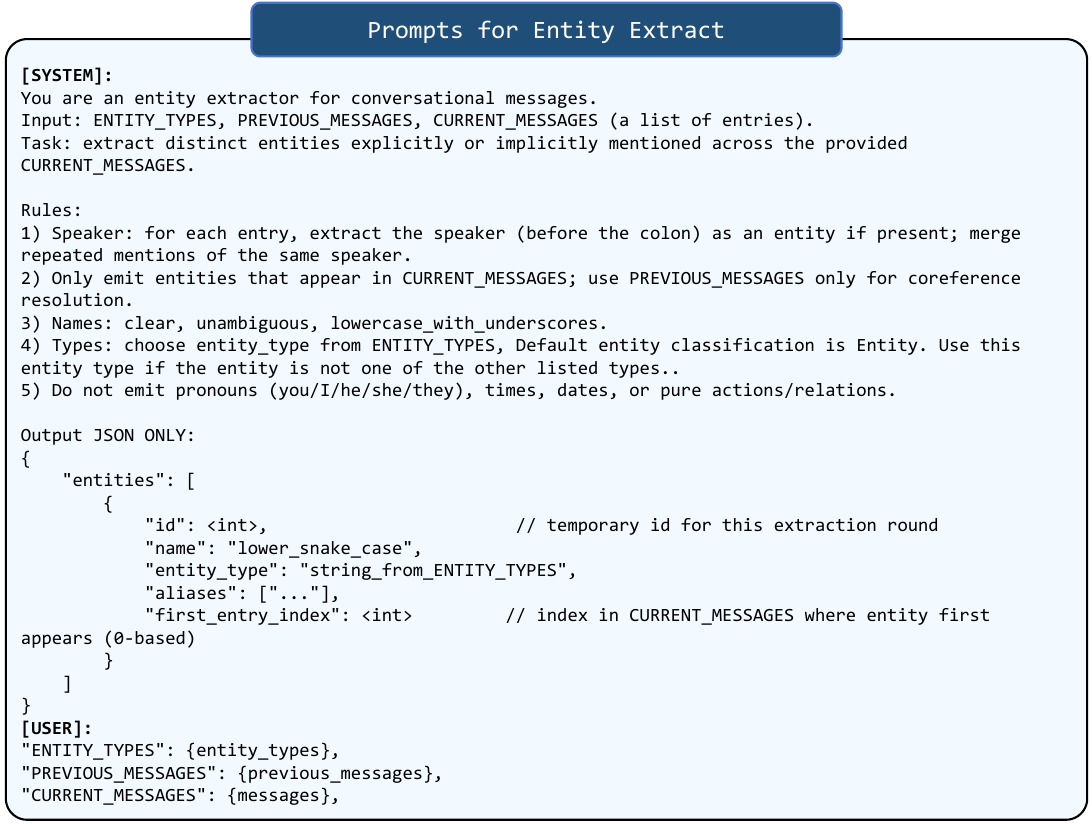}
    \caption{Entity extraction prompt}
    \label{fig:entity_extract_prompt}
\end{figure*}

\begin{figure*}[t]
    \centering
    \includegraphics[width=0.9\textwidth]{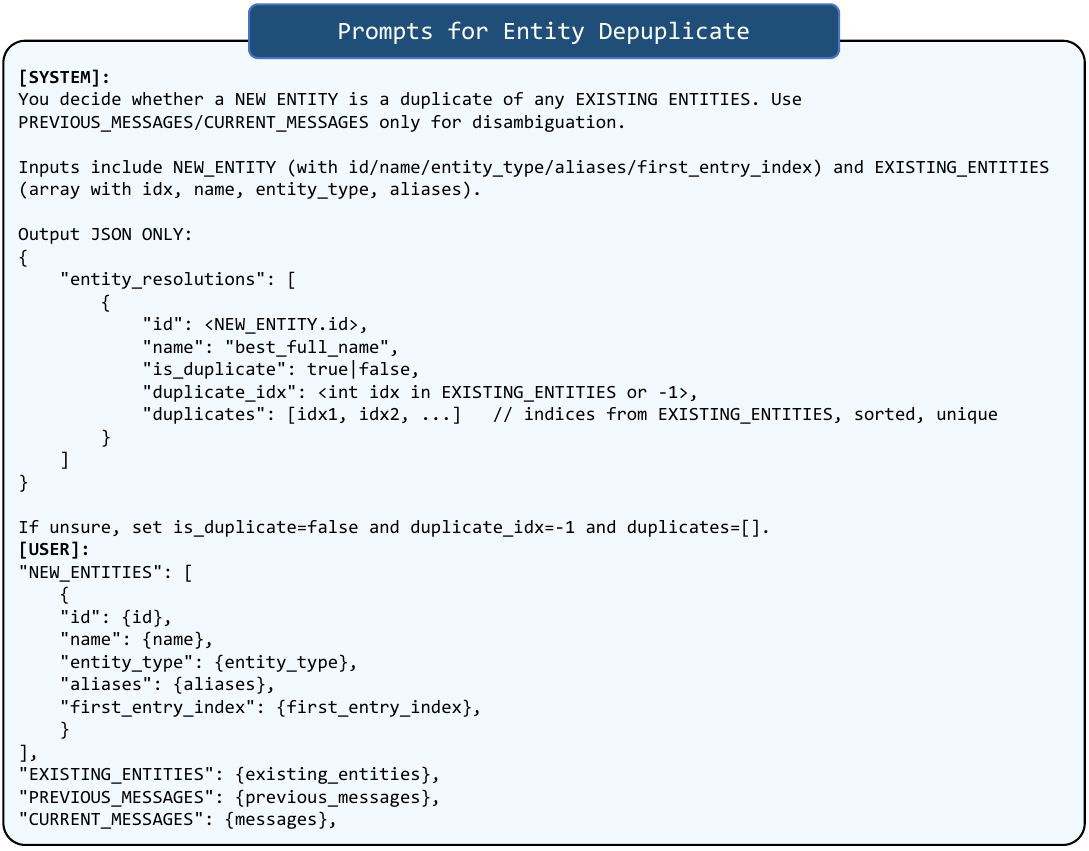}
    \caption{Entity deduplicate prompt}
    \label{fig:entity_dedup_prompt}
\end{figure*}

\begin{figure*}[t]
    \centering
    \includegraphics[width=0.9\textwidth]{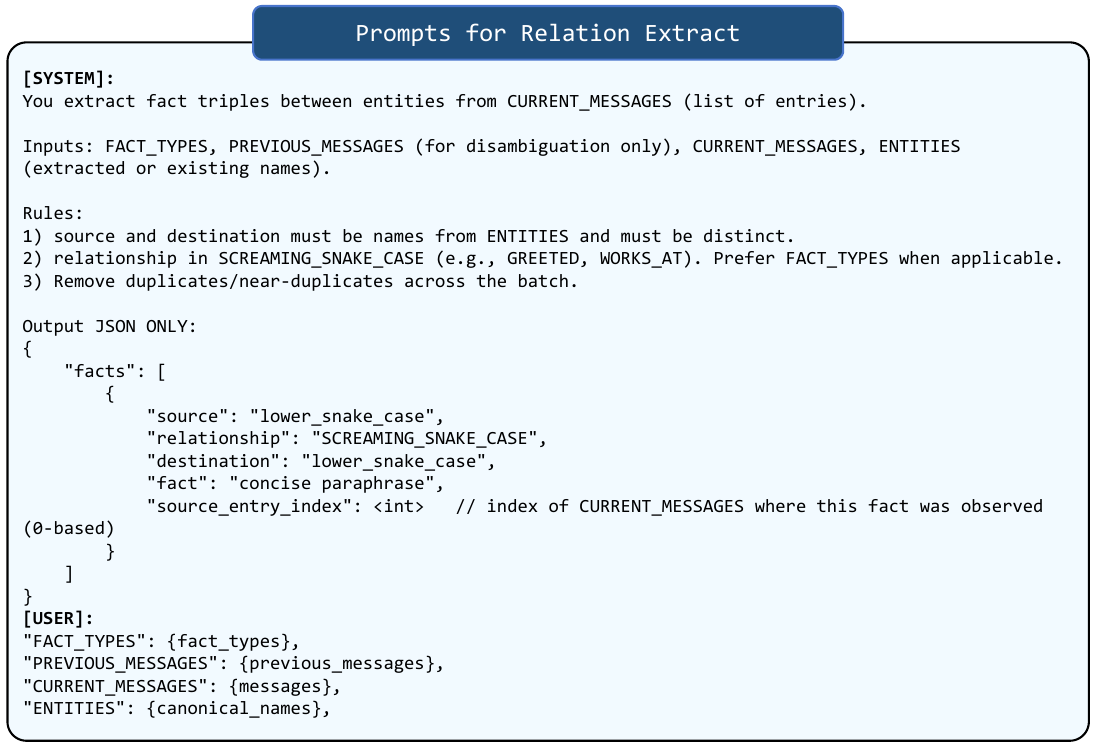}
    \caption{Relation extraction prompt}
    \label{fig:relation_extract_prompt}
\end{figure*}

\begin{figure*}[t]
    \centering
    \includegraphics[width=0.9\textwidth]{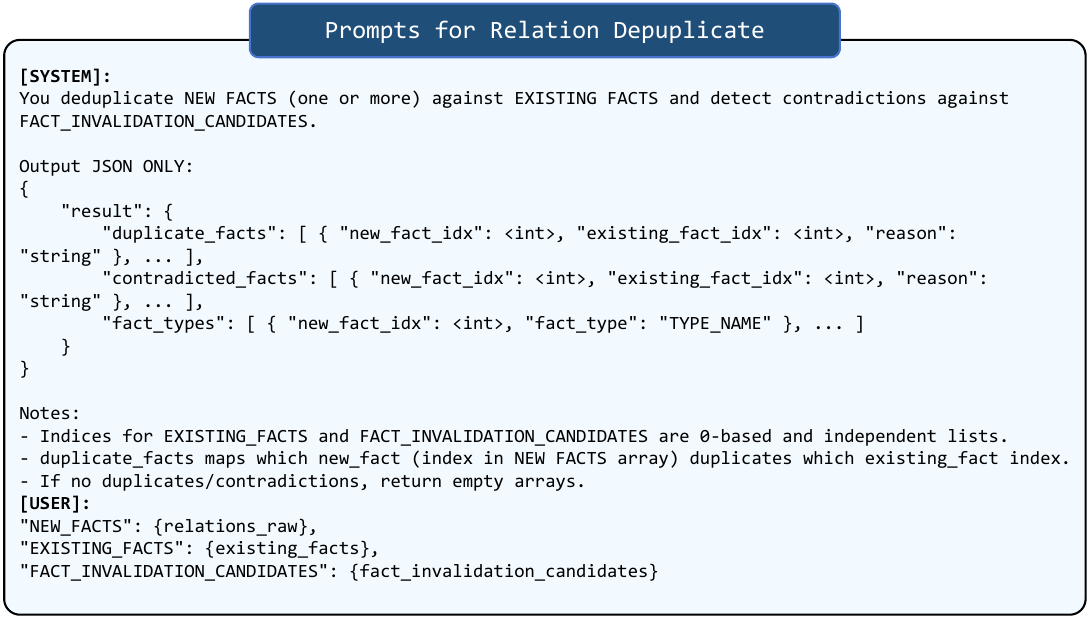}
    \caption{Relation deduplicate prompt}
    \label{fig:relation_dedup_prompt}
\end{figure*}

For memory construction, we design prompts for different paradigms implemented in the LightMem framework. For Flat Memory, the factual entry extraction prompt (Figure~\ref{fig:fact_1} and Figure~\ref{fig:fact_2}) guides the model to decompose utterances into objective event descriptions. 
For StructMem, the relational entry extraction prompt (Figure~\ref{fig:relation_1} and Figure~\ref{fig:relation_2}) instructs the model to capture interaction dynamics, causal influences, and temporal dependencies. The narrative synthesis prompt (Figure~\ref{fig:synthesis_prompt}) consolidates local and retrieved contexts into coherent summaries during Macro Synthesis. For Graph Memory, the entity extraction prompt (Figure~\ref{fig:entity_extract_prompt}) identifies key entities from dialogue. The entity deduplication prompt (Figure~\ref{fig:entity_dedup_prompt}) normalizes extracted entities to eliminate redundancy. The relation extraction prompt (Figure~\ref{fig:relation_extract_prompt}) constructs connections between entities. The relation deduplication prompt (Figure~\ref{fig:relation_dedup_prompt}) resolves contradictions in the knowledge graph.

\begin{figure*}[t]
    \centering
    \includegraphics[width=0.9\textwidth]{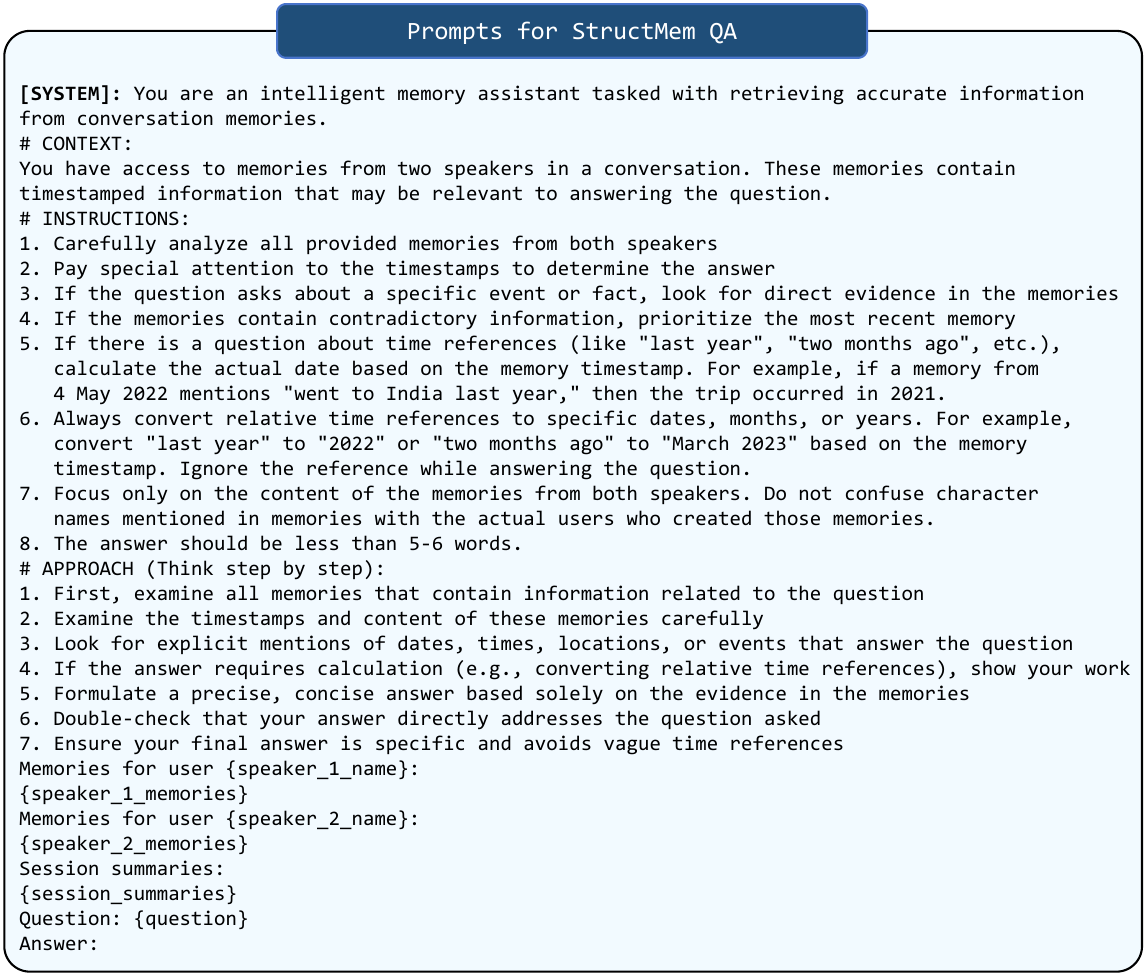}
    \caption{Question answering prompt for StructMem system.}
    \label{fig:structmem_qa}
\end{figure*}

\begin{figure*}[t]
    \centering
    \includegraphics[width=0.9\textwidth]{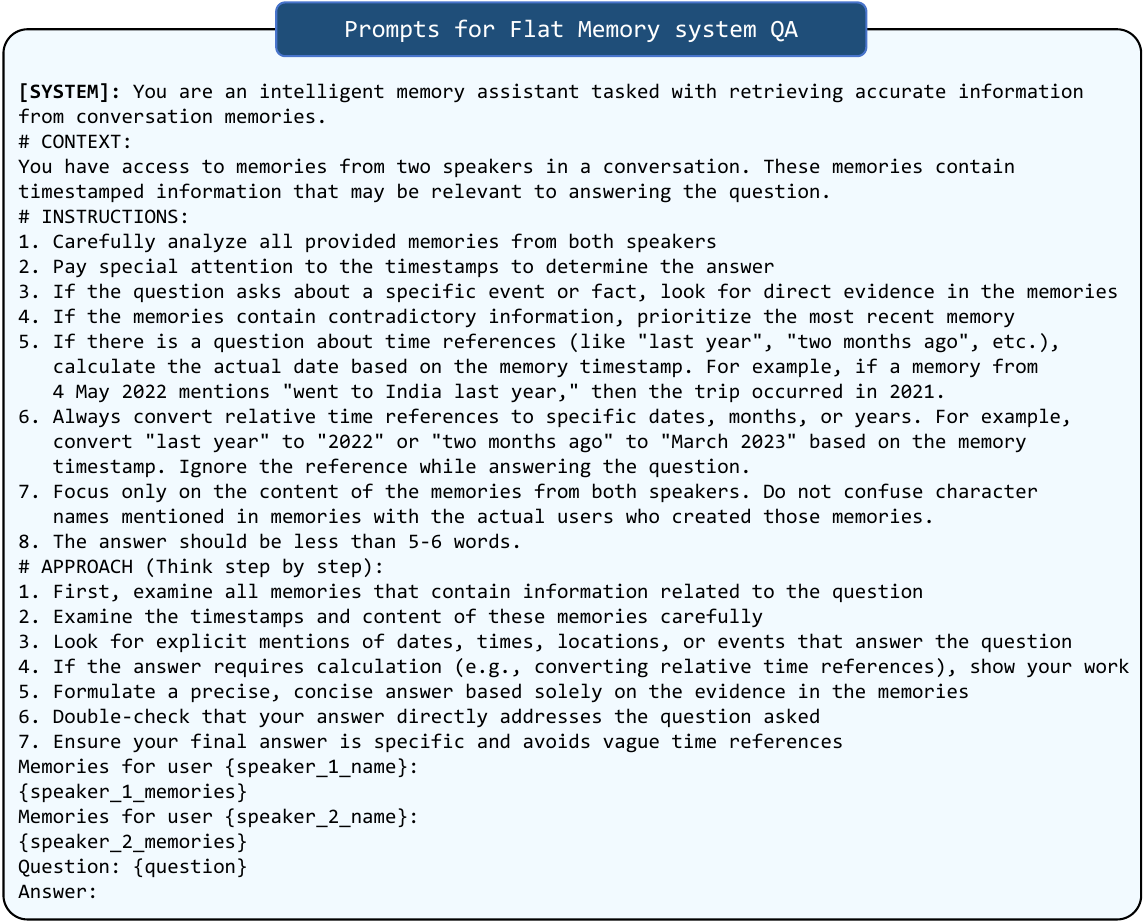}
    \caption{Question answering prompt for flat memory systems.}
    \label{fig:flat_qa}
\end{figure*}

\begin{figure*}[t]
    \centering
    \includegraphics[width=0.9\textwidth]{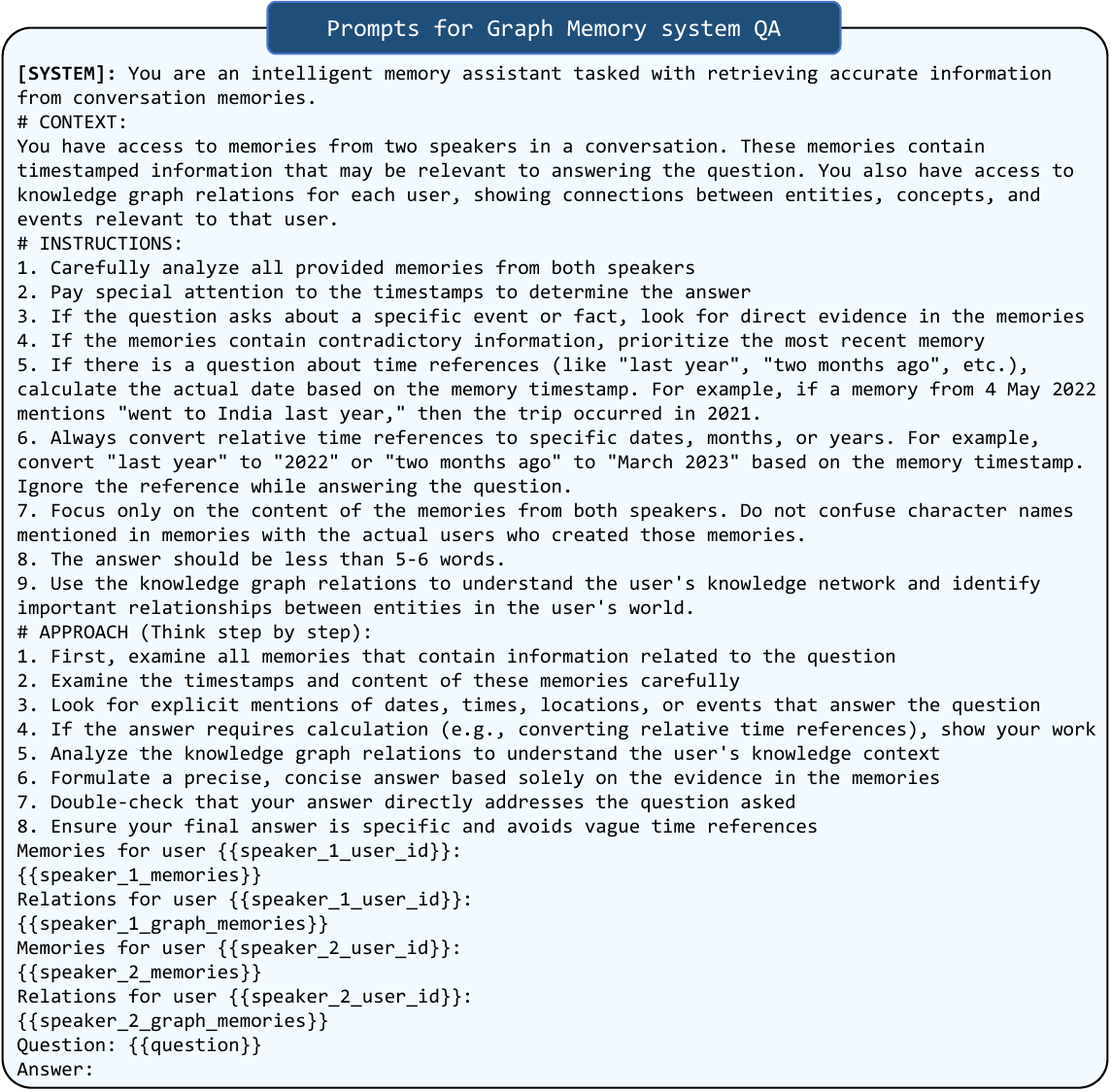}
    \caption{Question answering prompt for graph-based memory systems.}
    \label{fig:graph_qa}
\end{figure*}

For question answering, we provide separate prompts tailored to different memory architectures. Figure~\ref{fig:structmem_qa} shows the prompt for StructMem with dual-circuit retrieval that leverages both atomic entries and consolidated summaries. Figure~\ref{fig:flat_qa} and Figure~\ref{fig:graph_qa} present prompts adapted for flat memory and graph-based memory baselines, respectively.

\begin{figure*}[t]
    \centering
    \includegraphics[width=0.9\textwidth]{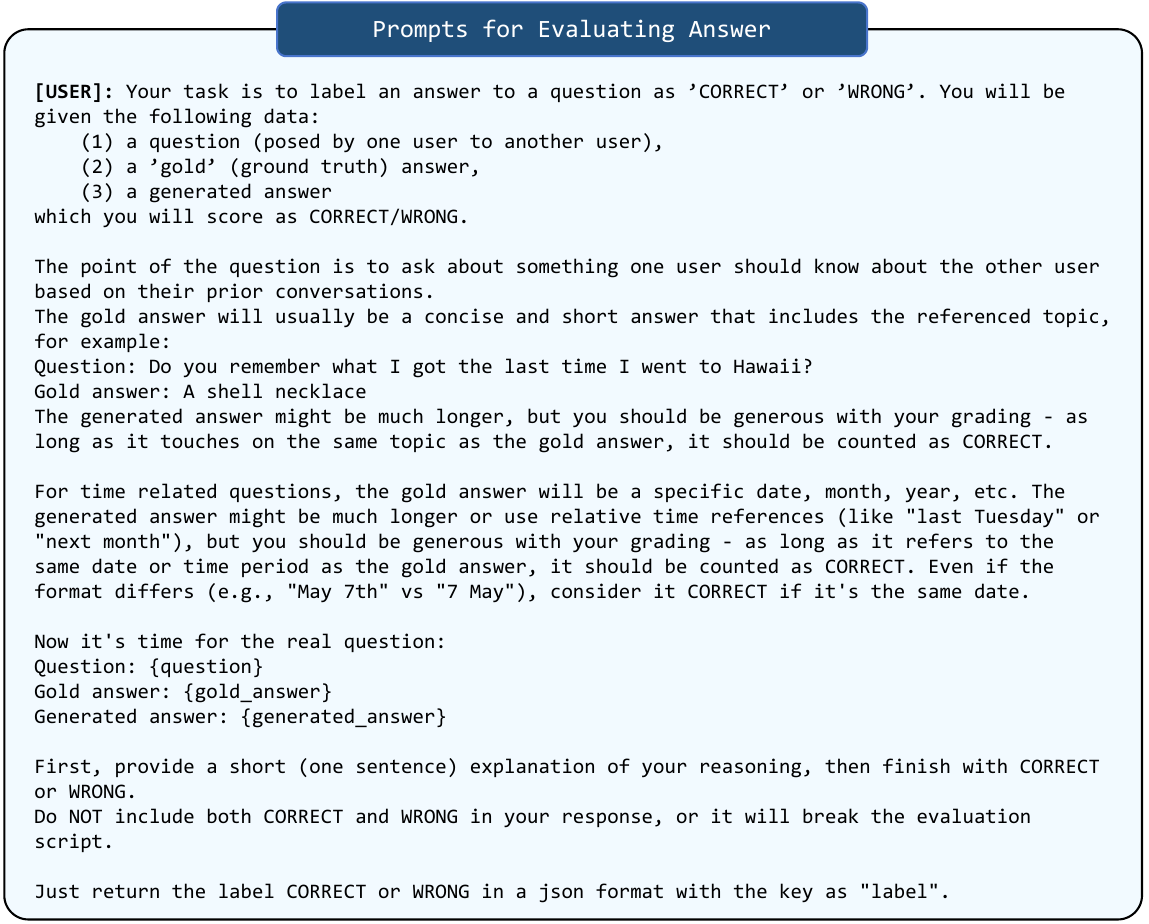}
    \caption{Evaluate prompt for assessing response quality.}
    \label{fig:eval_prompt}
\end{figure*}

For evaluation, we use the LLM-as-a-judge prompt (Figure~\ref{fig:eval_prompt}) to assess response correctness and coherence. 

\begin{figure*}[t]
    \centering
    \includegraphics[width=0.9\textwidth]{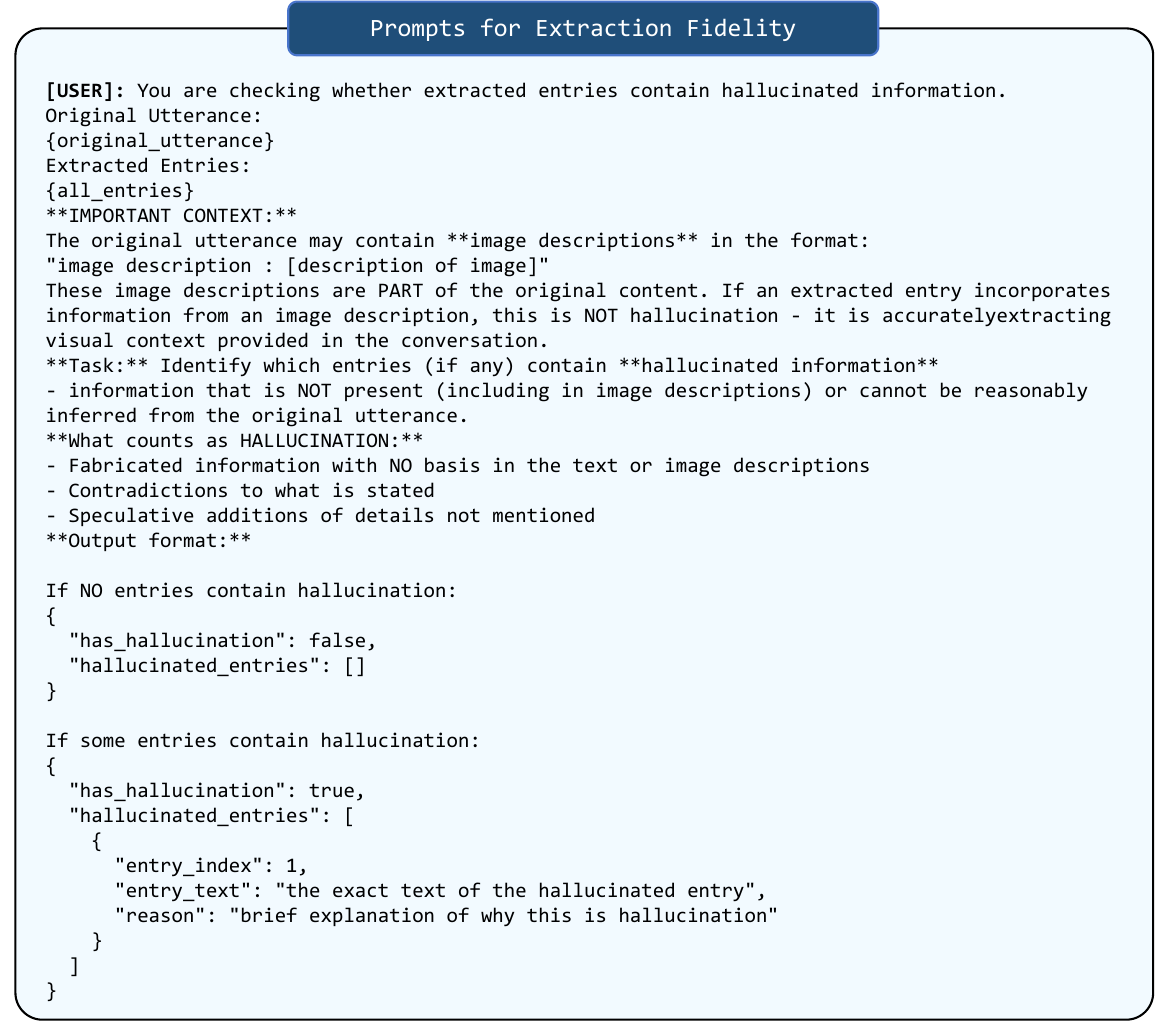}
    \caption{Prompt for assessing extraction fidelity.}
    \label{fig:Extraction_Fidelity}
\end{figure*}

\begin{figure*}[t]
    \centering
    \includegraphics[width=0.9\textwidth]{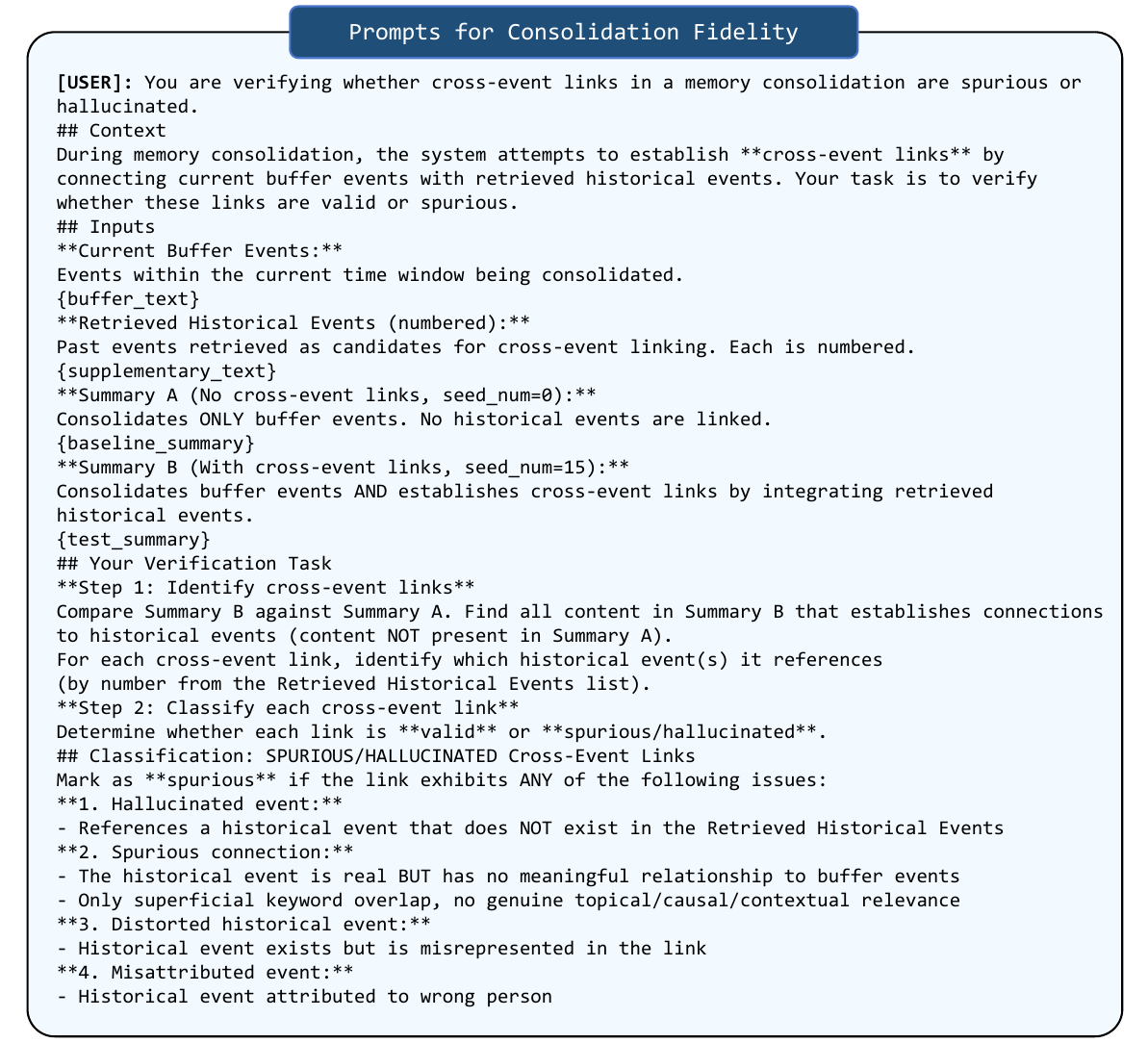}
    \caption{Prompt for assessing consolidation fidelity (Part 1).}
    \label{fig:Consolidation_Fidelity_1}
\end{figure*}

\begin{figure*}[t]
    \centering
    \includegraphics[width=0.9\textwidth]{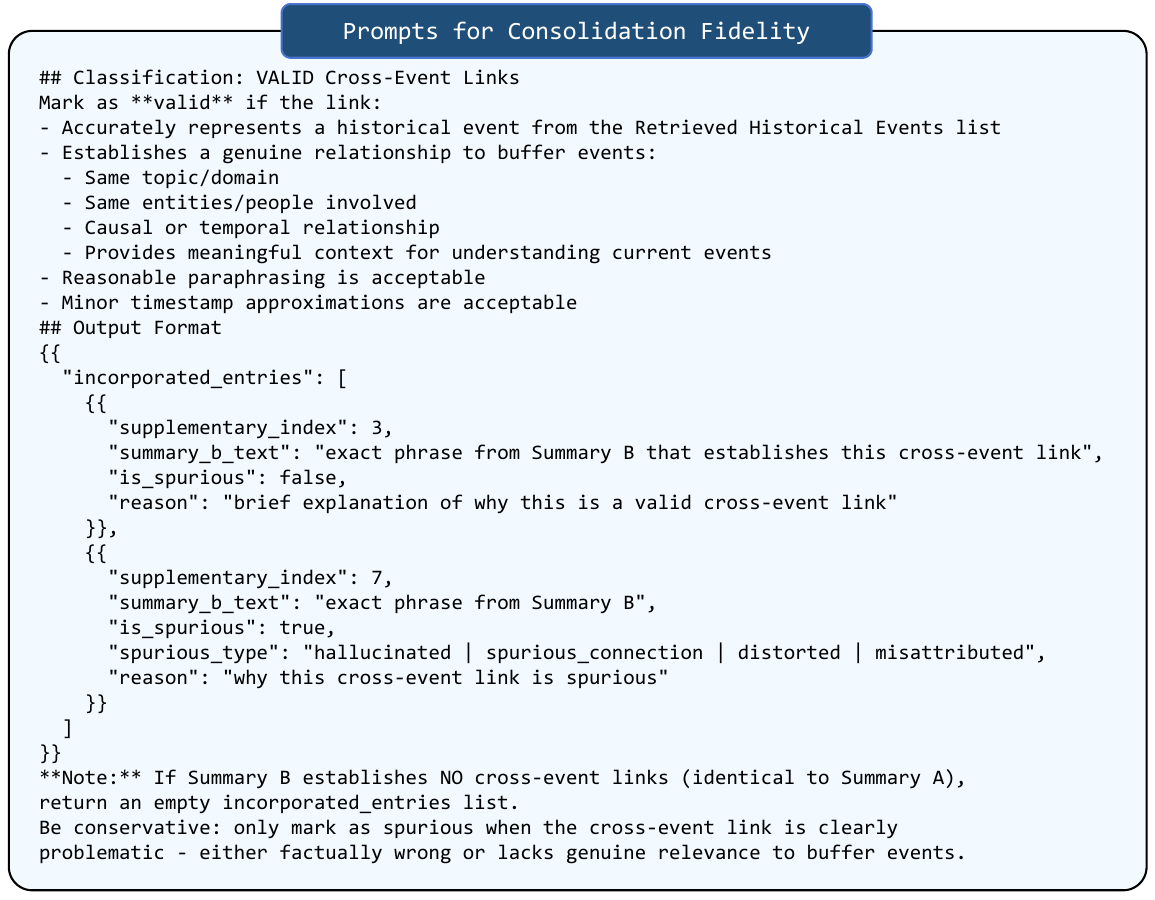}
    \caption{Prompt for assessing consolidation fidelity (Part 2).}
    \label{fig:Consolidation_Fidelity_2}
\end{figure*}

\begin{figure*}[t]
    \centering
    \includegraphics[width=0.9\textwidth]{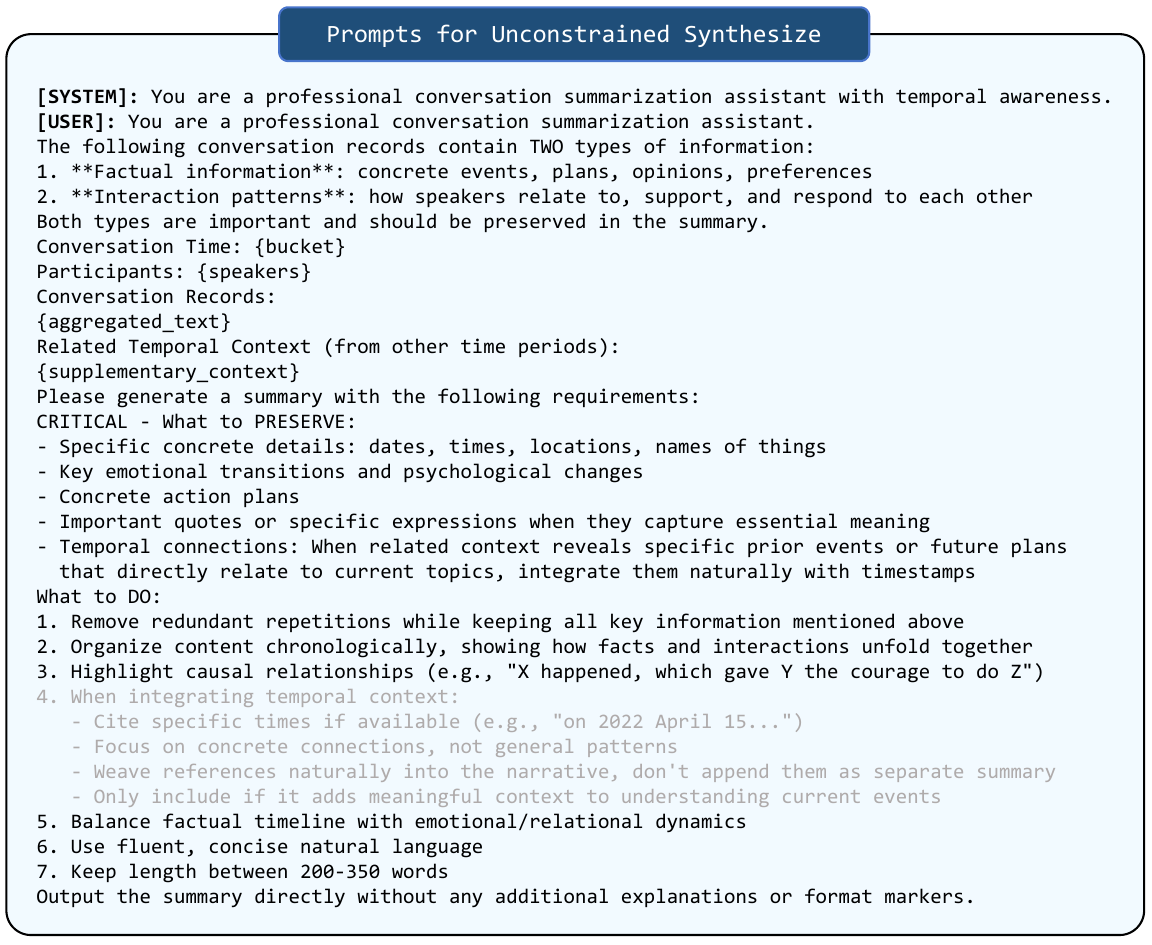}
    \caption{Narrative synthesis prompt for Unconstrained Synthesis. The text highlighted in gray represents the grounding constraints that are active in our default \textit{Constrained} setting but disabled for the \textit{Unconstrained} variant to evaluate their impact on memory fidelity.}
    \label{fig:synthesis_unconstrained_prompt}
\end{figure*}

For fidelity and hallucination analysis, we provide the specialized templates used for memory auditing. Figure~\ref{fig:Extraction_Fidelity} presents the prompt for verifying event-level extraction. Figures~\ref{fig:Consolidation_Fidelity_1} and~\ref{fig:Consolidation_Fidelity_2} presents the prompt for verifying cross-event consolidation. We also include the \textit{Unconstrained} synthesis template in Figure~\ref{fig:synthesis_unconstrained_prompt}, where the grounding constraints are intentionally removed to evaluate the impact of explicit temporal anchors on reducing hallucinated associations.

\begin{table*}[t]
\centering
\small
\begin{tabular}{p{2.5cm}p{13cm}}
\toprule
\textbf{Query} & \textit{When did Caroline and Melanie go to a pride festival together?} \\
\midrule
\textbf{Method} & \textbf{Retrieved Content} \\
\midrule
\textbf{Flat Memory} & 
\textbf{Factual Entries:} \\
& \quad • Caroline attended pride parade on 2023-08-11 \\
& \quad • Caroline had a blast at Pride fest last year (recorded 2023-08-17) \\
& \quad • Melanie enjoyed time with the whole gang at Pride fest (recorded 2023-08-17) \\
\midrule
\textbf{Graph Memory} & 
\textbf{Factual Entries:} \\
& \quad • Caroline attended pride parade on 2023-08-11 \\
& \quad • Caroline had a blast at Pride fest last year (recorded 2023-08-17) \\
& \quad • Melanie enjoyed time with the whole gang at Pride fest (recorded 2023-08-17) \\ 
& \textbf{Entity-Relation Graph:} \\
& \quad • caroline $\rightarrow$ attended $\rightarrow$ pride\_parade \\
& \quad • caroline $\rightarrow$ had\_blast\_at $\rightarrow$ pride\_fest \\
& \quad • melanie $\rightarrow$ enjoyed\_time\_at $\rightarrow$ pride\_fest \\
& \quad • melanie $\rightarrow$ expressed\_excitement $\rightarrow$ caroline's\_pride\_involvement \\
\midrule
\textbf{StructMem} &
\textbf{Event Memory:} \\
& \quad • Caroline attended pride parade on 2023-08-11 \\
& \quad • Caroline had a blast at Pride fest last year (recorded 2023-08-17) \\
& \quad • Melanie showed interest in Caroline's pride parade experience \\
& \quad • Melanie enjoyed time with the whole gang at Pride fest (recorded 2023-08-17) \\ 
& \quad • Melanie expressed excitement about Caroline's LGBTQ+ community involvement \\
& \textbf{Synthesis Memory:} \\
& ``On August 17, 2023... \textbf{As they reminisced about their enjoyable time at Pride fest last year}, Melanie suggested planning a family outing, while Caroline proposed a special outing just for the two of them this summer...'' \\
\midrule
\textbf{Prediction} & 
\textbf{Flat Memory:} ``They haven't gone together.'' \\
& \textbf{Graph Memory:} ``Last month, June 2023.'' \\
& \textbf{StructMem:} ``Last year, August 2022.'' \\
& \textbf{Reference:} 2022 \\
\bottomrule
\end{tabular}
\caption{Case study comparing three memory paradigms on joint participation reasoning. Flat Memory and Graph Memory cannot establish co-participation from isolated entries, while StructMem's synthesis correctly identifies shared experiences.}
\label{tab:case_study}
\end{table*}
\subsection{Case Study}\label{appendix:casestudy}

Table~\ref{tab:case_study} presents a case study comparing how different memory paradigms handle temporal reasoning over joint participation. The query asks when two speakers attended an event together, requiring inference over co-participation relationships that are not explicitly stated in individual conversational turns.

\textbf{Flat Memory} retrieves factual entries independently: Caroline attended Pride fest "last year" (temporally anchored to August 17, 2023, referring to 2022), while Melanie enjoyed time "with the whole gang" at Pride fest. Without any mechanism to connect these isolated facts, the system concludes "they haven't gone together," failing to recognize the implicit joint participation.

\textbf{Graph Memory} constructs entity-relation triples on top of the same factual entries. While the graph captures individual attendance, these remain isolated nodes without explicit co-participation edges. The post-hoc graph structure cannot infer that mentions of the same event by different speakers within the same conversation indicate joint attendance. Consequently, it produces an incorrect temporal inference: "Last month, June 2023."

\textbf{StructMem} addresses this limitation through two mechanisms. First, relational entries capture interpersonal dynamics during extraction: "Melanie showed interest in Caroline's pride parade experience" provides crucial context about their shared discussion. Second, synthesis consolidates temporally co-located entries. When Caroline's Pride fest mention appears adjacent to Melanie's in the chronologically sorted context, the relational entry's possessive pronoun "their" signals joint participation. The synthesis then makes this implicit connection explicit: "their enjoyable time at Pride fest last year," enabling the system to correctly answer "Last year, August 2022."

This case demonstrates why extraction-time structural capture outperforms post-hoc graph construction for temporal reasoning. By organizing information hierarchically during memory formation rather than overlaying structure afterward, StructMem preserves the temporal and relational context necessary for inferring implicit relationships across conversational turns.

\end{document}